\documentclass[times,twocolumn,final,5p]{elsarticle}
\usepackage{framed,multirow}

\usepackage{amssymb}
\usepackage{latexsym}

\usepackage{url}
\usepackage{xcolor}

\usepackage{hyperref}
\usepackage{bm}
\usepackage{url}
\usepackage{multirow}
\usepackage{ulem}
\usepackage{color}

\usepackage{amsmath,amssymb,amsfonts}
\usepackage{algorithmic}
\usepackage{graphicx}
\usepackage{textcomp}
\usepackage{bm}
\usepackage{url}
\usepackage{multirow}
\usepackage{ulem}
\usepackage{color}
\usepackage{pifont}
\usepackage{balance}
\usepackage{colortbl}

\newcommand{\cmark}{\ding{51}}
\newcommand{\xmark}{\ding{55}}
\definecolor{newcolor}{rgb}{.8,.349,.1}

\begin{document}
	
	%\verso{Shuwei Shao \textit{et~al.}}
	
	\begin{frontmatter}
		\title{Self-Supervised Monocular Depth and Ego-Motion Estimation in Endoscopy: Appearance Flow to the Rescue %\tnoteref{tnote1}}%
		}
		\tnotetext[t1]{This paper is an extensively extended version of ~\citet{shao2021self} presented at ICRA 2021.}
		\author[1]{Shuwei Shao}
		\cortext[cor1]{Corresponding authors}
		\author[1]{Zhongcai Pei}

		% \fntext[fn1]{This is author footnote for second author.
		\author[1]{Weihai Chen$^{*}$}
		\ead{whchen@buaa.edu.cn}
		\author[2]{Wentao Zhu}
		\author[1]{Xingming Wu}
		\author[3]{Dianmin Sun}
		\author[1]{Baochang Zhang$^{*}$}
		\ead{bczhang@buaa.edu.cn}
		%% Third author's email
		%\ead{author3@author.com}
		%\author[2]{Given-name4 \snm{Surname4}}
		
		\address[1]{School of Automation Science and Electrical Engineering, Beihang University, Beijing, China}
		\address[2]{Kuaishou Technology, USA}
		\address[3]{Shandong Cancer Hospital, Shandong University, Jinan, China}
		%\address[4]{Shenzhen Academy of Aerospace Technology, Shenzhen, China}
		
		%\received{-}
		%\finalform{-}
		%\accepted{-}
		%\availableonline{-}
		%\communicated{-}
		
		\normalem
	
		\begin{abstract}
			Recently, self-supervised learning technology has been applied to calculate depth and ego-motion from monocular videos, achieving remarkable performance in autonomous driving scenarios. One widely adopted assumption of depth and ego-motion self-supervised learning is that the image brightness remains constant within nearby frames. Unfortunately, the endoscopic scene does not meet this assumption because there are severe brightness fluctuations induced by illumination variations, non-Lambertian reflections and interreflections during data collection, and these brightness fluctuations inevitably deteriorate the depth and ego-motion estimation accuracy. In this work, we introduce a novel concept referred to as \textbf{\emph{appearance flow}} to address the brightness inconsistency problem. The appearance flow takes into consideration any variations in the brightness pattern and enables us to develop a \textbf{\emph{generalized dynamic image constraint}}. 
			Furthermore, we build a unified self-supervised framework to estimate monocular depth and ego-motion simultaneously in endoscopic scenes, which comprises a structure module, a motion module, an appearance module and a correspondence module, to accurately reconstruct the appearance and calibrate the image brightness. Extensive experiments are conducted on the SCARED dataset and EndoSLAM dataset, and the proposed unified framework exceeds other self-supervised approaches by a large margin. To validate our framework's generalization ability on different patients and 
			cameras, we train our model on SCARED but test it on the SERV-CT and Hamlyn datasets without any fine-tuning, and the superior results reveal its strong generalization ability. Code will be available at: \url{https://github.com/ShuweiShao/AF-SfMLearner}.
		\end{abstract}
		
		\begin{keyword}
		%% MSC codes here, in the form: \MSC code \sep code
		%% or \MSC[2008] code \sep code (2000 is the default)
		% \MSC 41A05\sep 41A10\sep 65D05\sep 65D17
		%% Keywords
		Self-supervised learning \sep  Monocular depth estimation \sep Ego-motion \sep Appearance flow \sep Brightness calibration  
	\end{keyword}
		
	\end{frontmatter}
	
	%\linenumbers
	
	%% main text
	\section{Introduction}
	\label{sec:introduction}

	Minimally invasive surgeries (MISs), \textit{e.g.}, laparoscopic procedures, have the potential advantages of less bleeding and lower infection rates than traditional open surgeries. However, MIS also has drawbacks, such as limited field of view, poor localization of endoscopy and lack of tactile feedback. The computer vision-based augmented reality (AR) navigation system is a feasible and low-cost solution to overcome these drawbacks, and it allows enhanced surgical visualization if the location of the endoscope is known. Generally, these systems require registering pre-operative data (\textit{e.g.}, computed tomography (CT) scans) to the intra-operative endoscopic videos~\citep{bernhardt2017status,chand2021challenge}. To ensure the reliability of an AR navigation system, a highly accurate registration process must be performed. The accuracy of video-CT registration algorithms primarily relies on the quality of intra-operative anatomical reconstructions from endoscopic videos. It is not trivial to acquire accurate and sufficiently dense reconstructions because of problems such as irradiance variations and the paucity of features.
	
\begin{figure}[!htb]
		\centering
		\includegraphics[width=1.0\linewidth]{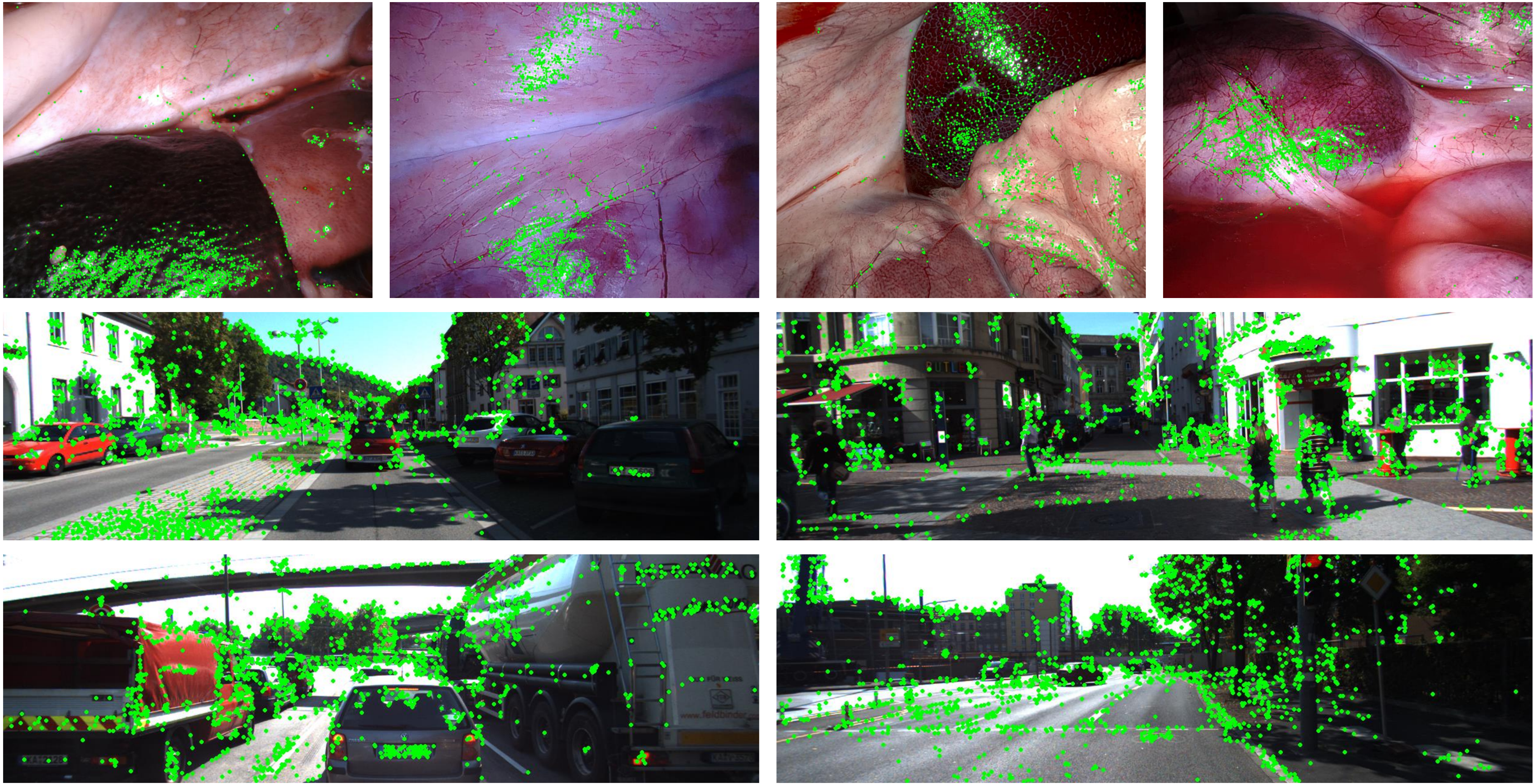}
		\caption{Key points detected by the scale-invariant feature transform (SIFT) algorithm~\citep{lowe2004distinctive} in an endoscopic scene and autonomous driving scenario, and denoted as green circles. The threshold for the key point extraction is set as $2000$.}	
		\label{Fig9}
	\end{figure}

	The reconstruction of 3D structures from monocular videos has been a longstanding research topic, and depth estimation is a necessary step to obtain reconstructions. Traditional multiview stereo methods, such as structure-from-motion (SfM)~\citep{leonard2018evaluation}, shape-from-shading (SfS)~\citep{ren2017shape} and simultaneous localization and mapping (SLAM)~\citep{chen2018slam}, are capable of reconstructing 3D structures in feature-rich scenes. Nevertheless, the sparse and unevenly distributed key points detected in endoscopic images (Fig.~\ref{Fig9}) can cause these multiview stereo methods to derive poor reconstructions.

	In recent years, deep learning-based methods have been proposed to predict depth maps from monocular videos. ~\citet{eigen2014depth},~\citet{xu2017multi},~\citet{cao2017estimating} and~\citet{fu2018deep} employed fully supervised convolutional neural networks and achieved outstanding depth estimates.
	Unfortunately, it is difficult to collect a large-scale and accurate RGB-D dataset based on endoscopic scenes due to the sensor noise and non-Lambertian reflection properties of tissues. In this work, we emphasize self-supervised monocular depth and ego-motion estimation~\citep{zhou2017unsupervised,zhan2018unsupervised,yin2018geonet,bian2019depth,godard2019digging,casser2019depth,ranjan2019competitive,luo2019every,fang2020towards,spencer2020defeat,yang2020d3vo}. Although considerable progress has been made in autonomous driving, there are still substantial challenges encountered when directly applying these methods to endoscopic videos.

\begin{figure*}[!htb]
		\centering
		\includegraphics[width=0.95\linewidth]{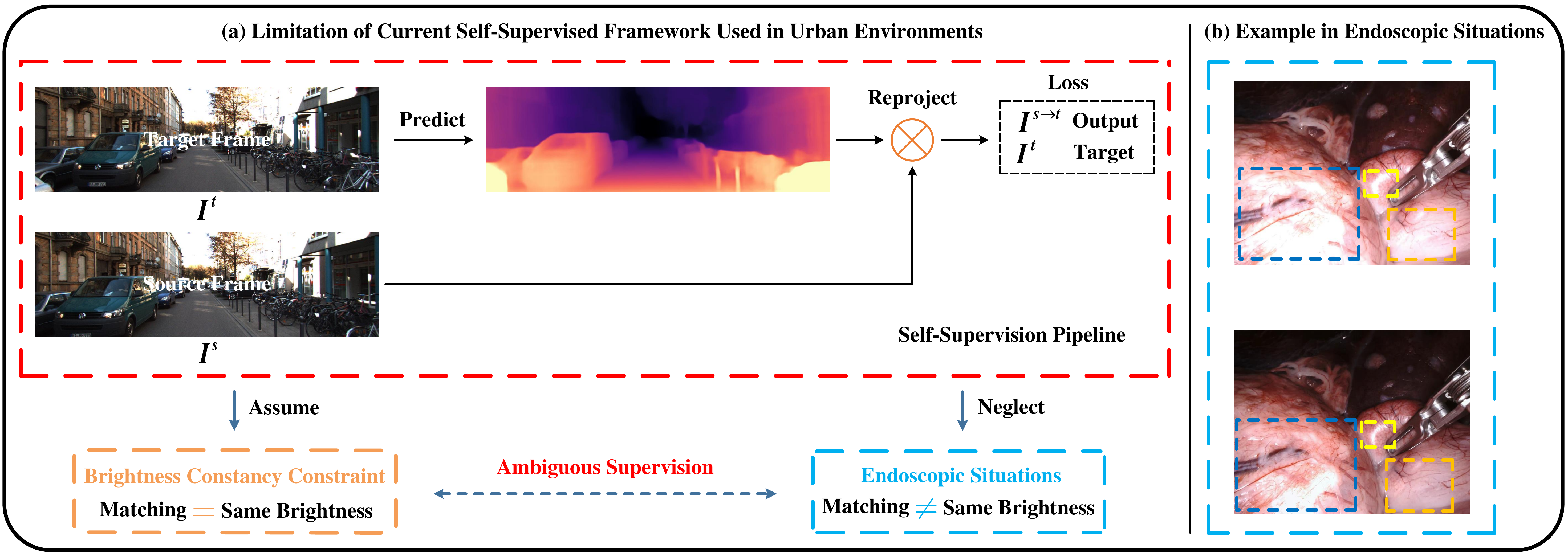}
		\caption{Illustration of the ambiguous supervision encountered when directly applying the self-supervised monocular depth and ego-motion estimation paradigm to endoscopic situations.}	
		\label{Fig1}
	\end{figure*}

	\normalem
	The self-supervised approaches are somewhat similar to the conventional SfM and make use of the disparity information contained in videos to supervise the deep networks. The core concept utilizes warping-based view synthesis via simultaneously estimated depth and ego-motion. The appearance difference between the target frame and the synthesized frame is delivered as a supervisory signal for the whole training phase. Depth and ego-motion self-supervised learning generally rests on a data constraint of \textbf{\emph{brightness constancy assumption}}\footnote{Basically, we refer to the pixel intensity as brightness.}~\citep{horn1981determining}. Specifically, the constraint assumes that intensity values of corresponding points remain (approximately) constant, which can hold roughly within the urban conditions. However, as Fig.~\ref{Fig1} shows, the brightness constancy assumption is invalid for endoscopy because severe variations are caused in illumination of the same anatomy from one frame to the next when the camera and light source move jointly. In addition, smooth tissues and organ fluids may contribute to strong non-Lambertian reflection and interreflection on their surfaces. Consequently, the appearance difference is susceptible to confusion by these brightness fluctuations, giving rise to ambiguous supervision in endoscopic scenes.

	Several methods have been developed to address the brightness inconsistency problem. \citet{liu2019dense} leveraged multiview stereo methods, \textit{e.g.}, SfM, to produce relatively robust sparse depths and camera poses. Thereafter, sparse depths are coupled with camera poses to supervise DepthNet. Nevertheless, it is laborious and time-consuming to preprocess large amounts of data with SfM. \citet{spencer2020defeat} proposed learning a dense visual representation in which the robust feature spaces are able to enhance the supervisory signal based on a warped feature coherency when the brightness constancy assumption expires. In contrast to urban environments, the overall texture observed in endoscopy is scarcer and more homogeneous, which is detrimental to visual representation learning. Even worse, this can introduce performance degradation up to incorrect depth estimates. \citet{yang2020d3vo} and \citet{bengisu2020quantitative} instead introduced affine transformers to align the target frame and its counterpart into similar brightness conditions. By applying the same transformation parameters to all pixel points in the target frame, the affine brightness transformer is inefficient when handling endoscopic images of complicated brightness variations, for example, induced by non-Lambertian reflection and interreflection occurring in the local regions~\citep{kharbat2008robust}.
	
\begin{figure*}[!htb]
		\centering
		\includegraphics[width=1.0\linewidth]{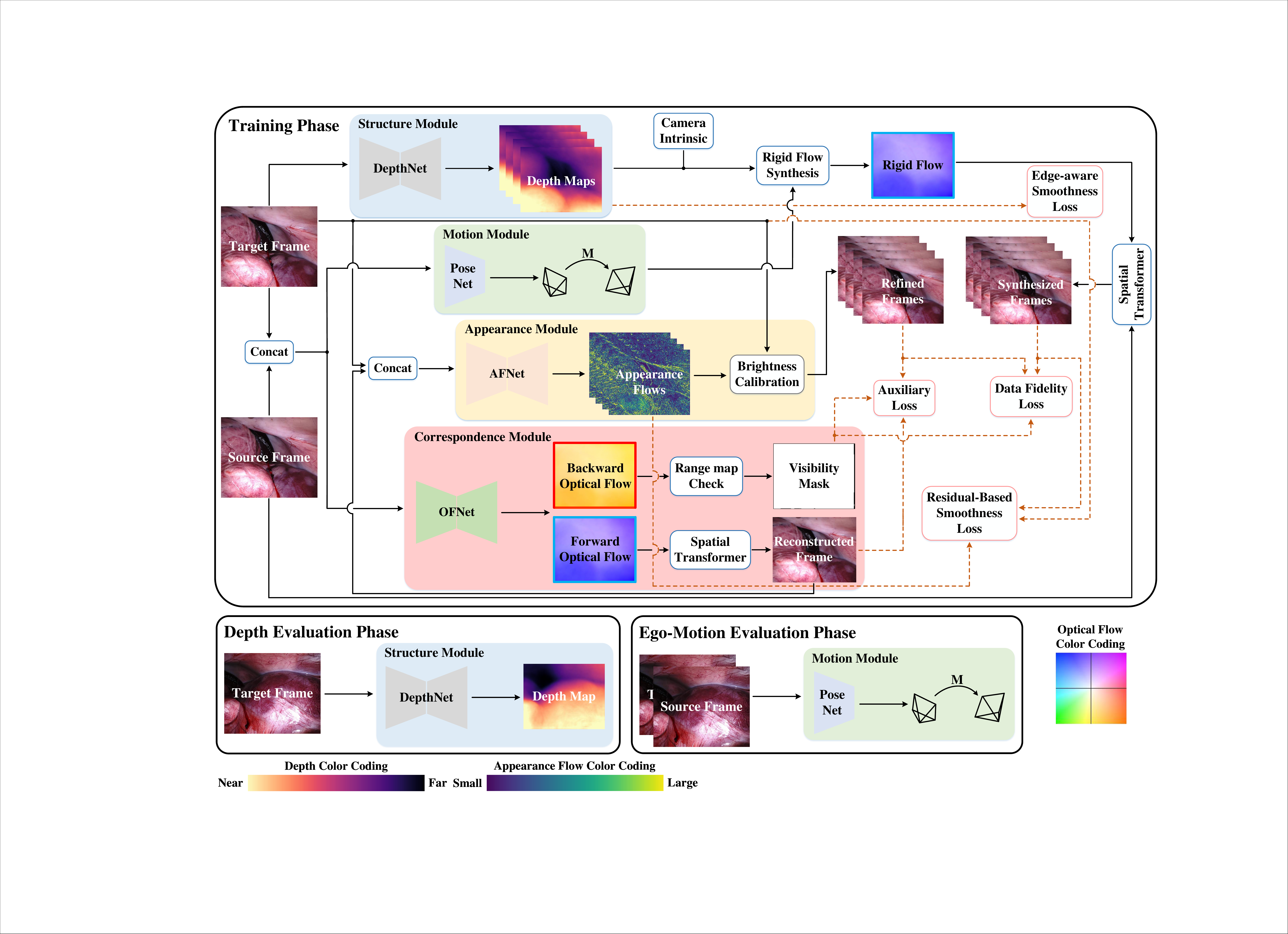}
		\caption{Overview of the unified self-supervised monocular depth and ego-motion estimation framework. Our network in the training phase (top) is composed of a structure module, a motion module, an appearance module and a correspondence module. Because of the spatial transformer's gradient locality~\citep{jaderberg2015spatial}, we use full-resolution multiscale depth prediction~\citep{godard2019digging}, appearance flow prediction and image reconstruction to achieve a robust training, and the number of scales is 4. The orange dashed arrows denote the data loss. During the evaluation phase (bottom), we use the trained DepthNet and PoseNet to estimate the dense depth maps and camera poses, respectively. Although DepthNet can predict depth maps at 4 scales, we only utilize the depth map generated at the highest resolution to evaluate the model.}	
		\label{Fig2}
	\end{figure*}

	In this work, we open up a new direction by utilizing \textbf{\emph{appearance flow}} between adjacent frames to address the brightness inconsistency problem. The concept of appearance flow is inspired by optical flow~\citep{horn1981determining,dosovitskiy2015flownet}, which is broadly applicable to video processing tasks~\citep{zhu2017deep}. The appearance flow incorporates effects from any underlying physical events that change the image brightness. Furthermore, the appearance flow induces a \textbf{\emph{generalized dynamic image constraint}} (GDIC), which includes both geometric and radiometric transformations, allowing a complete representation of interframe information and hence a relaxation of the brightness constancy assumption. The appearance flow field can vary between nearby pixel points. Except for isolated discontinuity boundaries caused by abrupt changes in illumination conditions and surface reflectance properties, these variations are expected to be smooth.
	With the above knowledge, we construct a unified self-supervised framework for the robust estimation of depth and ego-motion in endoscopic scenes, as shown in Fig.~\ref{Fig2}. The developed unified framework is composed of four parts: a structure module, a motion module, an appearance module and a correspondence module. In particular, the appearance module is used to predict appearance flow and calibrate the brightness condition. Moreover, the residual-based smoothness loss and auxiliary loss are designed to enforce the derived appearance flow with desired properties. Additionally, the correspondence module is employed to perform an automatic registration step, which exhibits significant efficacy in boosting the performance.  

	The main contributions of this work are summarized as follows:
	\begin{itemize}
		\item We introduce a novel concept of appearance flow that accounts for any variations in the brightness pattern. The appearance flow induces a generalized dynamic image constraint, in which geometric transformation can be combined with radiometric transformation to allow a complete information representation from one frame to the next. 
		\item We design several critical components, including an appearance module, a correspondence module, a residual-based smoothness loss and an auxiliary loss. On the basis of these components, we build a unified self-supervised framework that has high immunity against severe brightness fluctuations in endoscopic scenes.
		\item Detailed experiments and analysis demonstrate the effectiveness of our designed components in improving the depth and ego-motion estimation accuracy. The proposed unified framework outperforms previously competing self-supervised approaches on the SCARED dataset~\citep{allan2021stereo}, EndoSLAM dataset~\citep{bengisu2020quantitative}, SERV-CT dataset~\citep{edwards2020serv} and Hamlyn\footnote{\url{http://hamlyn.doc.ic.ac.uk/vision/}} dataset by a large margin. 
		
	\end{itemize}

	\section{Related work}
	In this section, we review relevant deep learning-based methods in the autonomous driving scenarios and endoscopic scenes.
	
	\subsection{Fully Supervised Depth Estimation.}
	As a pioneering work, \citet{eigen2014depth} proposed regressing depth from a single still image with multiscale networks and a scale-invariant loss. Since then, many follow-up works have continuously improved the depth estimation accuracy in various ways~\citep{liu2015learning,chen2016single,xu2017multi,cao2017estimating,fu2018deep,he2018learning,xu2018structured,repala2019dual, shao2021nenet}. \citet{xu2017multi} adopted a conditional random field (CRF) as a depth postprocessing module. \citet{cao2017estimating} and \citet{fu2018deep} treated the depth estimation as a classification problem and introduced more robust losses. These methods achieved excellent performance on datasets such as KITTI~\citep{geiger2012we} and Make3D~\citep{saxena2008make3d}. Nonetheless, fully supervised depth regression is challenging in endoscopic scenes due to the unavailability of a large-scale and accurate RGB-D dataset. \citet{visentini2017deep}, \citet{mahmood2018deep}, \citet{mahmood2018unsupervised} and \citet{chen2019slam} attempted to overcome this challenge through training on synthetic data. Their methods leveraged renderings of color images and depth maps to train a fully supervised DepthNet. In the evaluation phase, an appearance transform network was employed to convert real endoscopic images into simulation-like images or vice versa for depth prediction. However, the gap between real and synthetic domains is hard to make up by mimicking the appearance, which can lead to a decreased performance.
	
	\subsection{Self-Supervised Depth and Ego-Motion Estimation.}
	To eliminate the need for costly depth annotation, \citet{zhou2017unsupervised} developed a self-supervised framework that regards the depth learning problem as a warping-based view synthesis task. The framework contains a DepthNet and a separate PoseNet. To handle edge cases, such as object motion and occlusion/disocclusion, a predictive explainable mask was used.
	Inspired by this work, later works extended it by imposing extra geometric priors~\citep{yang2018lego,mahjourian2018unsupervised,chen2019self,bian2019depth}, introducing visual representation learning~\citep{zhan2018unsupervised,spencer2020defeat,shu2020feature}, and adding a self-attention mechanism~\citep{zhou2019unsupervised,johnston2020self}. Unfortunately, these methods may not generally be applicable to endoscopy due to the unique characteristics of minimally invasive surgery environments interframe brightness inconsistency, for example.
	
	\citet{turan2018unsupervised} proposed the first study of self-supervised depth and ego-motion estimation in endoscopic scenes, which is similar to \citet{zhou2017unsupervised}'s method. \citet{liu2019dense} employed sparse depths and camera poses generated from conventional SfM pipelines to establish supervision, where SfM was run as a preprocessing step. \citet{li2020unsupervised} considered the peak-signal-to-noise ratio (PSNR) as an additional optimization objective during the training phase. A recent work by \citet{bengisu2020quantitative} used an affine brightness transformer to enhance the photometric robustness, as well as a spatial attention module to dictate the PoseNet to emphasize highly textured regions.
	
	In contrast to the preceding methods, we introduce a novel concept called appearance flow and construct a robust unified self-supervised framework based on the designed components. In addition, our method is direct, and neither CT scans nor multiview stereo algorithms such as SfM are required. 
	
	\section{Methodology}
	\label{method}
	In this section, we first present preliminary knowledge regarding the self-supervised monocular depth and ego-motion estimation. Then, we investigate the brightness constancy assumption and introduce the appearance flow to develop a generalized dynamic image constraint. Finally, we build a unified self-supervised framework for robust depth and ego-motion estimation in endoscopic scenes.
	
	\subsection{Preliminaries}
	\label{pre}
	The framework of depth and ego-motion self-supervised learning generally involves two subnetworks: a DepthNet and a separate PoseNet. During the training phase, the key supervisory signal stems from warping-based view synthesis. Once the per-pixel depth values of the target frame are estimated, we back-project the pixel points on the image plane to a 3D camera space using known camera intrinsics. With the estimated ego-motion, the 3D point cloud can be projected onto another image plane. Given two frames, ${{{I}}^t}\left( {{\bm{{\rm{p}}}}} \right)$ and ${{I}^s}\left( {\bm{{\rm{p}}}} \right)$, the view synthesis is expressed as
	\begin{equation} 
		{h}\left( \bm{{\rm{p}}}^{s \to t}  \right) = {\left[ {\bf{K}\left| 0 \right.} \right]\bm{{\rm{M}}}^{t \to s} \left[ \begin{array}{c}
				\bm{{\rm{D}}}^{t}{\bm{{\rm{K}}}^{ - 1}}{h}\left( {\bm{{\rm{p}}}^{t}} \right)\\
				1
			\end{array} \right]}, \label{eq1}
	\end{equation}	
	where ${h}\left( \bm{{\rm{p}}}^{s \to t}  \right)$ and ${h}\left( \bm{{\rm{p}}}^{t}  \right)$ are the homogeneous pixel coordinates in source view $s$ and target view $t$, respectively, $\bm{{\rm K}}$ is the camera intrinsic, ${\bm{{\rm M}}^{t \to s}}$ is the ego-motion from $t$ to $s$, and ${\bm{{\rm D}}^t}\left( \bm{{\rm{p}}} \right)$ is the depth map of target frame ${{{I}}^t}\left( {{\bm{{\rm{p}}}}} \right)$. Then, we can obtain a rigid flow between $t$ and $s$:
	\begin{equation} {\bm{{\rm{F}}}_\delta^{t \to s}}\left( {\bm{{\rm{p}}}} \right) = {\bm{{\rm{p}}}^{s \to t}} - {\bm{{\rm{p}}}^t}, \label{eq2} \end{equation}  
	and synthesize the frame ${{{I}}^{s \to t}}\left( {{\bm{{\rm{p}}}}} \right)$ via a differentiable inverse warping operation (spatial transformer)~\citep{jaderberg2015spatial}. 
	The appearance difference between target frame ${{{I}}^t}\left( {{\bm{{\rm{p}}}}} \right)$ and synthesized frame ${{I}^{s \to t}}\left( {\bm{{\rm{p}}}} \right)$ supervises the entire training pipeline.

    The common practice~\citep{godard2017unsupervised,bian2019depth,spencer2020defeat} to assess the appearance difference is to use a combination of L1 loss and structural similarities (SSIM) term~\citep{wang2004image}
		\begin{equation} \resizebox{0.91\hsize}{!}{$\Phi \left( {{I^t},{I^{s \to t}}} \right) = {\alpha}\displaystyle{\frac{1 - {\rm{SSIM}}\left( {{I^t},{I^{s \to t}}} \right)}{2}} + \left( {1 - \alpha } \right){\left| {{I^t} - {I^{s \to t}}} \right|}.$} \end{equation} The SSIM term is relatively robust to the data noises in the real-world scenario, as it considers the statistics of an image patch, \textit{e.g. }, the mean and variance.
	
	\subsection{Brightness Constancy Assumption}
	One widely adopted assumption of self-supervised depth and ego-motion estimation is the interframe brightness constancy, which poses a restriction that the rigid flow describes the displacements of pixel points with constant image brightness. For a sequence of images $I\left( {{\bm{{\rm{p}}}},\tau } \right)$, where ${\bm{{\rm{p}}}} = {\left[ {u,v} \right]^T}$ and $\tau$ denotes time, this is equivalent to determining a route $\bm{{\rm p}}\left( \tau \right)$ along which the image brightness stays invariant,
	\begin{equation}
		I\left( {\bm{{\rm p}}\left( \tau \right),\tau } \right) = c, \label{eq4}
	\end{equation}
	where $c$ stands for brightness. Applying the temporal derivative to both sides of Eq. \ref{eq4}, we arrive at
	\begin{equation}
		\frac{{{\text{d}}I}}{{{\text{d}}\tau }} = {\left( {\nabla I} \right)^T}\bm{{\rm{f}}} + {I_\tau } = 0, \label{eq5}
	\end{equation}
	where ${\nabla I} = {\left[ {\partial I{\text{/}}\partial u ,\partial I{\text{/}}\partial v } \right]^T}$ stands for the spatial gradient field, ${I_\tau } = \partial I{\text{/}}\partial \tau $ stands for the brightness change rate at a fixed point in the image, and $\bm{{\rm{f}}} = {\left[ {{\text{d}}u{\text{/d}}\tau ,{\text{d}}v{\text{/d}}\tau } \right]^T}$.
	We multiply Eq. \ref{eq5} by the time interval $\delta \tau$ to obtain
	\begin{equation}
		{\left( {\nabla I} \right)^T}{\bm{{\rm{F}}}_\delta }   + {I_\tau }\delta \tau = 0.
	\end{equation}
	According to Eq. \ref{eq1} and Eq. \ref{eq2}, using depth $\bm{{\rm{D}}}$ and ego-motion $\bm{{\rm{M}}}$ in place of $\bm{{\rm{F}}}_\delta$, we can obtain a constraint equation with respect to $\bm{{\rm{D}}}$ and $\bm{{\rm{M}}}$ under the brightness constancy assumption
	\begin{equation} {\left( {\nabla I} \right)^T}\underbrace {{\left( {\left[ {\bm{{\rm{K}}}\left| 0 \right.} \right]\bm{{\rm{M}}}\left[ \begin{array}{c}
						\bm{{\rm{D}}}{\bm{{\rm{K}}}^{ - 1}}h\left(  \bm{{\rm{p}}}  \right)\\
						1
					\end{array} \right]} \right)_ \times }}_{\text{\small{G}}}{\kern 1pt} + {I_\tau }\delta \tau = {\left( {\nabla I} \right)^T}\bm{{\rm{p}}},	\label{eq7}
	\end{equation}
	where ${\left(  \cdot  \right)_ \times }$ stands for the nonhomogeneous pixel coordinate and G stands for the geometric transformation.
	
	  \begin{figure}[!htb]
		\centering
		\includegraphics[width=1.0\linewidth]{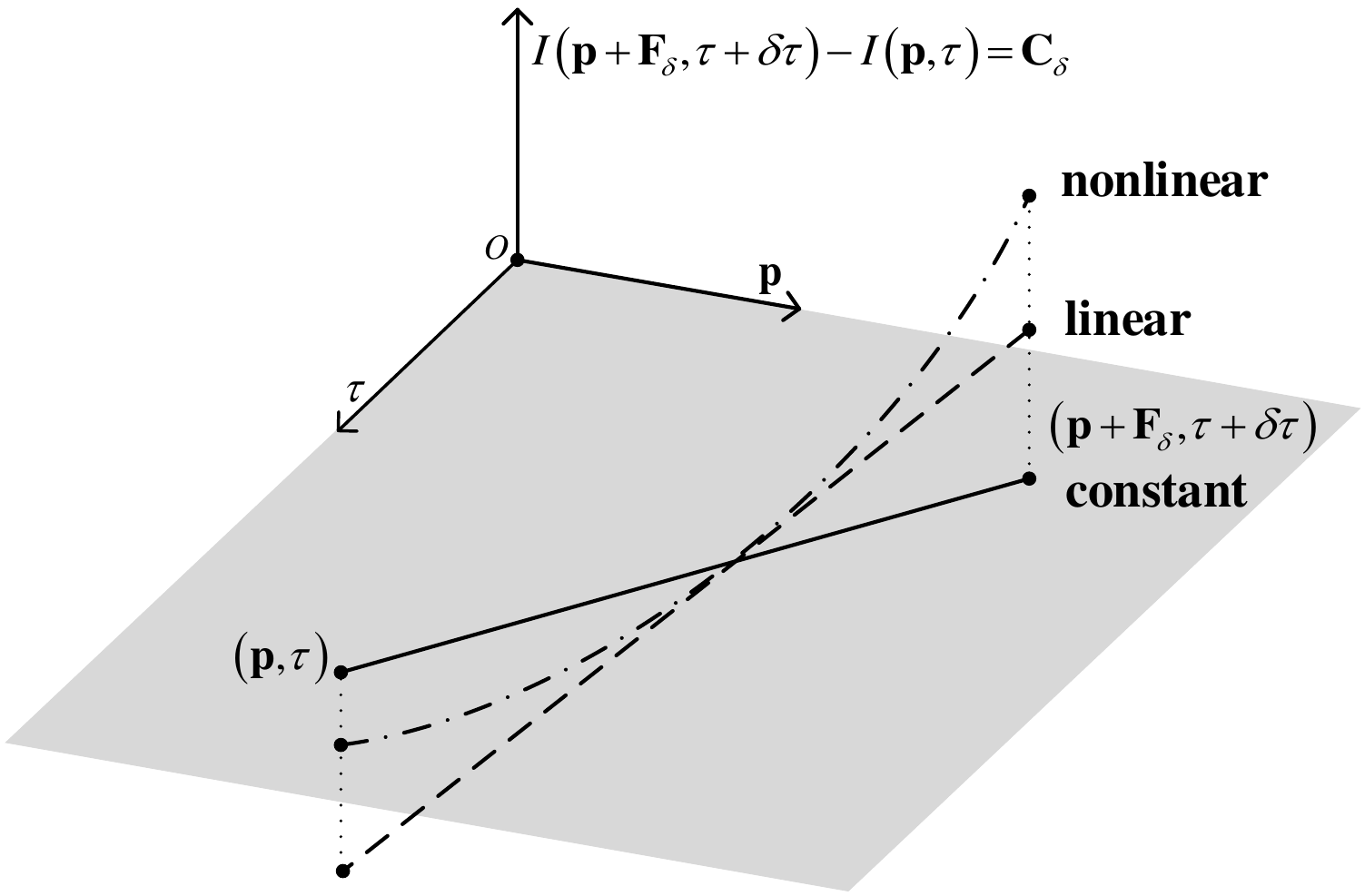}
		\caption{Illustration of the generalized dynamic image constraint that enables image brightness to vary (linear and nonlinear) from frame to frame. By contrast, the brightness constancy assumption restricts the path $\bm{{\rm p}}\left( \tau \right)$ to the gray plane ($I\left( {{\bf{p}} + {{\bf{F}}_\delta },\tau  + \delta \tau } \right) - I\left( {{\bf{p}},\tau } \right) = 0$).}
		\label{Fig10}
	\end{figure}

	In the minimally invasive surgery scenes, several aggregated physical events can lead to severe image brightness fluctuations, including light source motion, non-Lambertian reflection and interreflection. 
	As a result, the appearance difference computed based on the brightness constancy assumption can be an extremely biased supervisory signal, which inevitably deteriorates the depth and ego-motion estimation accuracy. To properly resolve the problems brought by the irradiance variations of scene surfaces, image brightness constraints that involve radiometric transformation are required.
	
	\subsection{Appearance Flow}
	\label{AF}
	\subsubsection{Generalized Dynamic Image Constraint}
	In this work, we develop a \textbf{\emph{generalized dynamic image constraint}} to describe both geometric and radiometric transformations in an image sequence
	\begin{equation} {I_{syn}}\left( {\bm{{\rm{p}}},\tau} \right) = I\left( {\bm{{\rm{p}}} + {{{\bf{F}}_\delta }},\tau  + \delta \tau } \right) = I\left( {\bm{{\rm{p}}},\tau } \right) + {{{\bf{C}}_\delta }}\left( {\bm{{\rm{p}}},\tau } \right), \label{eq9}
	\end{equation}
	where ${I_{syn}}\left( {\bm{{\rm{p}}},\tau} \right)$ is the synthesized image from time $\tau  + \delta \tau$ to time $\tau$ via view synthesis, and the last term ${{{\bf{C}}_\delta }}\left( {\bm{{\rm{p}}},\tau } \right)$, which we call \textbf{\emph{appearance flow}}, describes the radiometric transformation (R) in a sequence of images. The superiority of the generalized dynamic image constraint against the brightness constancy assumption is demonstrated in Fig.~\ref{Fig10}.
	
	We use the first-order Taylor series expansion of formula $I\left( {\bm{{\rm{p}}} + {\bf{F}}_\delta,\tau  + \delta \tau } \right)$ and acquire
	\begin{equation} I\left( {\bm{{\rm{p}}} + {\bf{F}}_\delta,\tau  + \delta \tau } \right) = I\left( {\bm{{\rm{p}}},\tau } \right) + {\left( {\nabla I} \right)^T}{\bf{F}}_\delta + {I_\tau }\delta \tau. \label{eq10}
	\end{equation}
	Then, substituting Eq. \ref{eq10} into Eq. \ref{eq9} and simplifying, we obtain
	\begin{equation} {\left( {\nabla I} \right)^T}{\bf{F}}_\delta + {I_\tau }\delta \tau = {{{\bf{C}}_\delta }}.
	\end{equation}
	We replace $\bm{{\rm{F}}}_\delta$ with Eq. \ref{eq1} and Eq. \ref{eq2}, and can achieve a generalized constraint equation with respect to depth $\bf{D}$, ego-motion $\bf{M}$ and appearance flow ${{{\bf{C}}_\delta }}$
	\begin{equation}
		\resizebox{0.89\hsize}{!}{$
			\begin{array}{l}
				{\left( {\nabla I} \right)^T}\underbrace {{\left( {\left[ {\bm{{\rm{K}}}\left| 0 \right.} \right]\bm{{\rm{M}}}\left[ \begin{array}{c}
								\bm{{\rm{D}}}{\bm{{\rm{K}}}^{ - 1}}h\left(  \bm{{\rm{p}}}  \right)\\
								1
							\end{array} \right]} \right)_ \times }}_{\text{\small{G}}} + {I_\tau }\delta \tau = {\left( {\nabla I} \right)^T}\bm{{\rm{p}}} + \underbrace {{{{\bf{C}}_\delta }}}_\text{\small{R}}
			\end{array}.$} \label{eq12}
	\end{equation}
	In the special case of ${{{\bf{C}}_\delta }} = 0$, Eq. \ref{eq12} reduces to the constraint equation Eq. \ref{eq7}. It is worth emphasizing the necessity of appearance flow for self-supervised monocular depth and ego-motion estimation in endoscopic scenes:
	
	\begin{itemize}
		\item The relative motion of camera and scene surfaces in 3D space, which induces a 2D pixel motion field on the image plane through projection geometry, is only one particular physical event.
		\item Other physical events, \textit{e.g.}, motion of the light source, non-Lambertian reflection and interreflection lead to complicated variations in the brightness pattern. As stated above, severe brightness fluctuations can force erroneous estimates of depth and ego-motion.
		\item Geometric transformation, when combined with radiometric transformation, represents complete information in an image sequence, allowing perfect reconstruction from one image $I\left( {\bm{{\rm{p}}} + {{{\bf{F}}_\delta }},\tau  + \delta \tau } \right)$ to the image $I\left( {\bm{{\rm{p}}},\tau } \right)$. In this case, the supervisory signal delivered by the appearance difference is precise and reliable.
	\end{itemize}
	\subsubsection{The Ill-Posed Problem}  
	Self-supervised monocular depth and ego-motion estimation in light of the brightness constancy assumption is a potentially unconstrained problem because only one constraint equation for two unknowns is accessible (Eq.\ref{eq7}). Hence, there exist many possible solutions to correctly synthesize the novel view when given two frames. This ambiguity is referred to as the ``ill-posed problem". The previous self-supervised approaches typically resolve the solution ambiguity by imposing smoothness on the depth structure, for example,~\cite{godard2017unsupervised}. In our generalized dynamic image constraint, the indetermination of solutions is even worse: unknowns in Eq.\ref{eq12} are three. Since appearance flow is unconstrained, the depth and ego-motion can be fully arbitrary, with the appearance flow altering in such a way as to ensure that the constraint equation Eq.\ref{eq12} is satisfied. Arriving at solutions that are consistent with physical events necessitates enforcing appropriate regularization. ~\\
	
	\noindent \textbf{Smoothness constraint prior.} Occluding boundaries, such as those incurred by the occlusion of surgical instruments and tissues and anatomical edges, can give rise to discontinuities in the depth map. In addition, discontinuities in the appearance flow field may occur if the illumination conditions or the surface reflectance characteristics change abruptly as the camera and light source move in endoscopic scenes. In the absence of occluding boundaries, anatomical edges, abrupt variations in illumination conditions and reflectance characteristics, the depth map and appearance flow field are expected to be smooth. Given these facts, we require that depth, ego-motion and appearance flow conform to the generalized dynamic image constraint and that the depth map and appearance flow field vary smoothly.
	
	\subsection{A Unified Self-Supervised Framework for the Robust Estimation of Depth and Ego-Motion in Endoscopic Scenes}
	We build a novel unified self-supervised framework to estimate monocular depth and ego-motion simultaneously, as illustrated in Fig.~\ref{Fig2}. The unified framework includes a structure module, a motion module, an appearance module and a correspondence module to accurately reconstruct the appearance and calibrate the brightness. The structure module is a monodepth estimator that converts a target frame into a dense depth map. The motion module functions as a 6DOF ego-motion estimator, which takes two adjacent frames as input and outputs a relative pose parameterized by Euler angles and a translation vector. The appearance module is used to predict appearance flow and align the brightness condition through a brightness calibration procedure
	\begin{equation} 
		{{I}^t}\left( {\bm{{\rm{p}}}} \right) + {{{\bf{C}}_\delta }}\left( {\bm{{\rm{p}}}} \right) \Leftrightarrow {{I}^s}\left( {\bm{{\rm{p}}}} \right).
	\end{equation}
	The correspondence module is intended to carry out an automatic registration step. 
	A simple example clarifies the utility of the registration step. Consider the following scenario: a stationary scene, a fixed camera and a moving light source. For video frames taken in this scenario, only irradiance variations occur. Therefore, the brightness changes between adjacent frames can be acquired using a simple subtraction operation. Similarly, the registration step minimizes the motion component and enables irradiance variations to become prominent between adjacent frames, which is beneficial to the derivation of appearance flow and shows significant effectiveness in improving the depth and ego-motion accuracy. It should be noted that the registered frames are used as the input of AFNet, and the input of PoseNet is the raw data. In addition, the correspondence module also yields a visibility mask that filters out occluded or out-of-view pixel points.
		
	The optimization objectives for training the proposed framework are divided into two parts: a data fidelity term ${\cal D}\left( {\bm{{\rm{p}}}} \right)$ and a Tikhonov regularizer ${\cal R}\left( {\bm{{\rm{p}}}} \right)$ \begin{equation} 	{{\cal L}_{self}} = {\cal D}\left( {\bm{{\rm{p}}}} \right) + \kappa {\cal R}\left( {\bm{{\rm{p}}}} \right), \label{16} \end{equation}
	where $\kappa$ is a hyperparameter used to balance data fidelity and regularization, thus constraining the solution space. The ${\cal D}\left( {\bm{{\rm{p}}}} \right)$ term is defined as
	\begin{equation}
		{\cal D}\left( {\bm{{\rm{p}}}} \right) = \sum\limits_{\bm{{\rm{p}}}} {{\bm{{\rm{V}}}}\left( \bm{{\rm{p}}} \right) * \Phi \left( {{{I}^{s \to t}}\left( \bm{{\rm{p}}} \right)}, {{I}^t}\left( \bm{{\rm{p}}} \right) + {{{\bf{C}}_\delta }}\left( {\bm{{\rm{p}}}} \right)\right)}, 
	\end{equation} where ${{{I}^{s \to t}}\left( \bm{{\rm{p}}} \right)}$ is synthesized by the rigid flow and spatial transformer. In our setting, ${{I}^s}\left( {\bm{{\rm{p}}}} \right)$ comprises two nearby views of ${{I}^t}\left( {\bm{{\rm{p}}}} \right)$, i.e., ${{I}^s}\left( {\bm{{\rm{p}}}} \right) \in \left\{ {{{I}^{t - 1}\left( {\bm{{\rm{p}}}} \right)},{{I}^{t + 1}\left( {\bm{{\rm{p}}}} \right)}} \right\}$, and ${{\bm{{\rm{V}}}}\left( \bm{{\rm{p}}} \right)}$ is a visibility mask and can be produced through the following range map~\citep{wang2018occlusion} check:
	\begin{equation}
		\resizebox{0.89\hsize}{!}{$
			\begin{array}{l}
				\bm{{\rm{R}}}\left( {u,v} \right) = \sum\limits_{i = 1}^W {\sum\limits_{j = 1}^H {\max \left( {0,1 - \left| {u - \left( {i + \bm{{\rm{F}}}_{bf}^u\left( {i,j} \right)} \right)} \right|} \right)} } \\
				{\kern 1pt} {\kern 1pt} {\kern 1pt} {\kern 1pt} {\kern 1pt} {\kern 1pt} {\kern 1pt} {\kern 1pt} {\kern 1pt} {\kern 1pt} {\kern 1pt} {\kern 1pt} {\kern 1pt} {\kern 1pt} {\kern 1pt} {\kern 1pt} {\kern 1pt} {\kern 1pt} {\kern 1pt} {\kern 1pt} {\kern 1pt} {\kern 1pt} {\kern 1pt} {\kern 1pt}{\kern 1pt} {\kern 1pt} {\kern 1pt} {\kern 1pt} {\kern 1pt} {\kern 1pt} {\kern 1pt} {\kern 1pt} \cdot \max \left( {0,1 - \left| {v - \left( {j + \bm{{\rm{F}}}_{bf}^v\left( {i,j} \right)} \right)} \right|} \right)
			\end{array}$}
	\end{equation}
	and
	\begin{equation}
		{\bm{{\rm{V}}}}\left( {u,v} \right) = \bm{{\rm{R}}}\left( {u,v} \right) > 0.95,
	\end{equation}
	where $\bm{{\rm{R}}}\left( {u,v} \right)$ is the range map that indicates the possibility of nonoccluded regions, $\left( {W,H} \right)$ stands for the image width and height, $\left( {\bm{{\rm{F}}}_{bf}^u,\bm{{\rm{F}}}_{bf}^v} \right)$ stands for the horizontal and vertical components of backward optical flow, and 0.95 is a threshold\footnote{We select the threshold by referring to the setting in~\citep{wang2018occlusion} and finetune it on the SCARED dataset.} to obtain the visibility mask $\bm{{\rm{V}}}\left( {u,v} \right)$.
	
	The Tihkonov regularizer ${\cal R}\left( {\bm{{\rm{p}}}} \right)$ consists of three terms
	\begin{equation} {\cal R}\left( {\bm{{\rm{p}}}} \right) = \lambda_1{{\cal L}_{rs}} + \lambda_2{{\cal L}_{ax}} + \lambda_3{{\cal L}_{es}}. 
	\end{equation}
	Details of ${{\cal L}_{rs}}$, ${{\cal L}_{ax}}$ and ${{\cal L}_{es}}$ are described below.~\\
	
	\noindent \textbf{Residual-based smoothness loss.} ${{\cal L}_{rs}}$ is defined to encourage the smoothness property of the appearance flow field, which penalizes the first-order gradients %of appearance flowBased on
	\begin{equation}
		{{\cal L}_{rs}} = \sum\limits_{\bm{{\rm{p}}}} {\left| { {\nabla {\bf{C}_\delta }\left( \bf{p} \right)}} \right|}.
	\end{equation}
	Furthermore, the gradients of the residual error are used to emphasize the regions with sharp brightness variations
	\begin{equation}
		{{\cal L}_{rs}} = \sum\limits_{\bm{{\rm{p}}}} {\left| { {\nabla {\bf{C}_\delta }\left( \bf{p} \right)}} \right|} * {e^{ - \nabla \left| {{I^t}\left( \bf{p} \right) - {I^{s \to t}}\left( \bf{p} \right)} \right|}}.
	\end{equation}
	\noindent \textbf{Auxiliary loss.} ${{\cal L}_{ax}}$ is defined to provide an auxiliary supervisory signal for the AFNet. 
	Mathematically,
	\begin{equation}
		{{\cal L}_{ax}} = \sum\limits_{\bm{{\rm{p}}}} {{\bm{{\rm{V}}}}\left( \bm{{\rm{p}}} \right) * \Phi \left( {{{I}^{s \to t}}\left( \bm{{\rm{p}}} \right)}, {{I}^t}\left( \bm{{\rm{p}}} \right) + {{{\bf{C}}_\delta }}\left( {\bm{{\rm{p}}}} \right)\right)}, 
	\end{equation}
	where ${{{I}^{s \to t}}\left( \bm{{\rm{p}}} \right)}$ is reconstructed by the optical flow and spatial transformer. The use of auxiliary loss to regularize network training is not new, \citep{zhao2017pyramid} and \citep{Sun_2019_CVPR}, for example. Our novelty is in using an optical flow network to improve an appearance flow network.~\\
	
	\noindent \textbf{Edge-aware smoothness loss.} ${{\cal L}_{es}}$ is applied to enforce the smoothness property of the depth map
	\begin{equation}
		{{\cal L}_{es}} = \sum\limits_p {\left| { {\nabla {\bf{D} }\left( \bf{p} \right)}} \right|} * {e^{ - \nabla \left| {{I^t}\left( \bf{p} \right)} \right|}},
	\end{equation}
	which is proposed in ~\citep{godard2017unsupervised}.~\\
		
	\noindent \textbf{Network architectures.} This work focuses on dealing with the brightness inconsistency problem, and thus our network design is mainly consistent with existing self-supervised methods. Here is a brief description of architectures used in this work.
	
	The DepthNet employs an encoder-decoder structure with skip connections, in which ResNet-18~\citep{he2016deep} with the fully connected layer removed is used as the encoder, and the design of the decoder is the same as Monodepth2~\citep{godard2019digging}. For PoseNet, we adopt the model of Monodepth2~\citep{godard2019digging}, which is a lightweight network. The AFNet is based on a similar structure to DepthNet. In the encoding stage, a concatenated image pair passes through layers of convolutional blocks with a stride of 2, resulting in a five-level feature pyramid. Then, skip connections propagate the pyramid features into the decoding stage. We alternate between the nearest upsampling layer, concatenating feature maps, $3 \times 3$ convolution with ELU activation, and estimation layer up to the highest resolution. The OFNet maintains the same architecture as AFNet, except for the estimation layer. The detailed decoder architectures of AFNet and OFNet are shown in Table~\ref{table1}. Note that our unified framework is agnostic to the specific selection of each subnetwork, and other choices are feasible. 
	
	\begin{table}[htb]
		\caption{The decoder architecture of AFNet and OFNet. AF: Appearance Flow; OF: Optical Flow. All layers are convolutional layers with the kernel size of $3 \times 3$ and stride of 1, and $\uparrow$ represents the $2 \times 2$ nearest-neighbor upsampling. } 
		\begin{center}
			\smallskip
			\renewcommand{\arraystretch}{1.2}
			\resizebox{1.0\columnwidth}{!}{\begin{tabular}{c c c c c}
					\hline
					\textbf{layer} & \textbf{channels} & \textbf{resolution scale} &  \textbf{input} & \textbf{activation}\\
					\hline
					\hline
					upconv5 & 256 & 32 & econv5 & ELU \\
					iconv5 & 256 & 16 & $\uparrow$upconv5, econv4 & ELU \\
					upconv4 & 128 & 16 & iconv5 & ELU \\
					iconv4 & 128 & 8 & $\uparrow$upconv4, econv3 & ELU \\
					AF4 & 3 & 8 & iconv4 & Tanh \\
					OF4 & 2 & 8 & iconv4 & Linear \\
					upconv3 & 64 & 8 & iconv4 & ELU \\
					iconv3 & 64 & 4 & $\uparrow$upconv3, econv2 & ELU \\
					AF3 & 3 & 4 & iconv3 & Tanh \\
					OF3 & 2 & 4 & iconv3 & Linear \\
					upconv2 & 32 & 4 & iconv3 & ELU \\
					iconv2 & 32 & 2 & $\uparrow$upconv2, econv1 & ELU \\
					AF2 & 3 & 2 & iconv2 & Tanh \\
					OF2 & 2 & 2 & iconv2 & Linear \\
					upconv1 & 16 & 2 & iconv2 & ELU \\
					iconv1 & 16 & 1 & $\uparrow$upconv1 & ELU \\
					AF1 & 3 & 1 & iconv1 & Tanh \\
					OF1 & 2 & 1 & iconv1 & Linear \\
					\hline
			\end{tabular}}
		\end{center}
		\label{table1}
	\end{table}
	\begin{figure*}[!htb]
		\centering
		\includegraphics[width=1.0\linewidth]{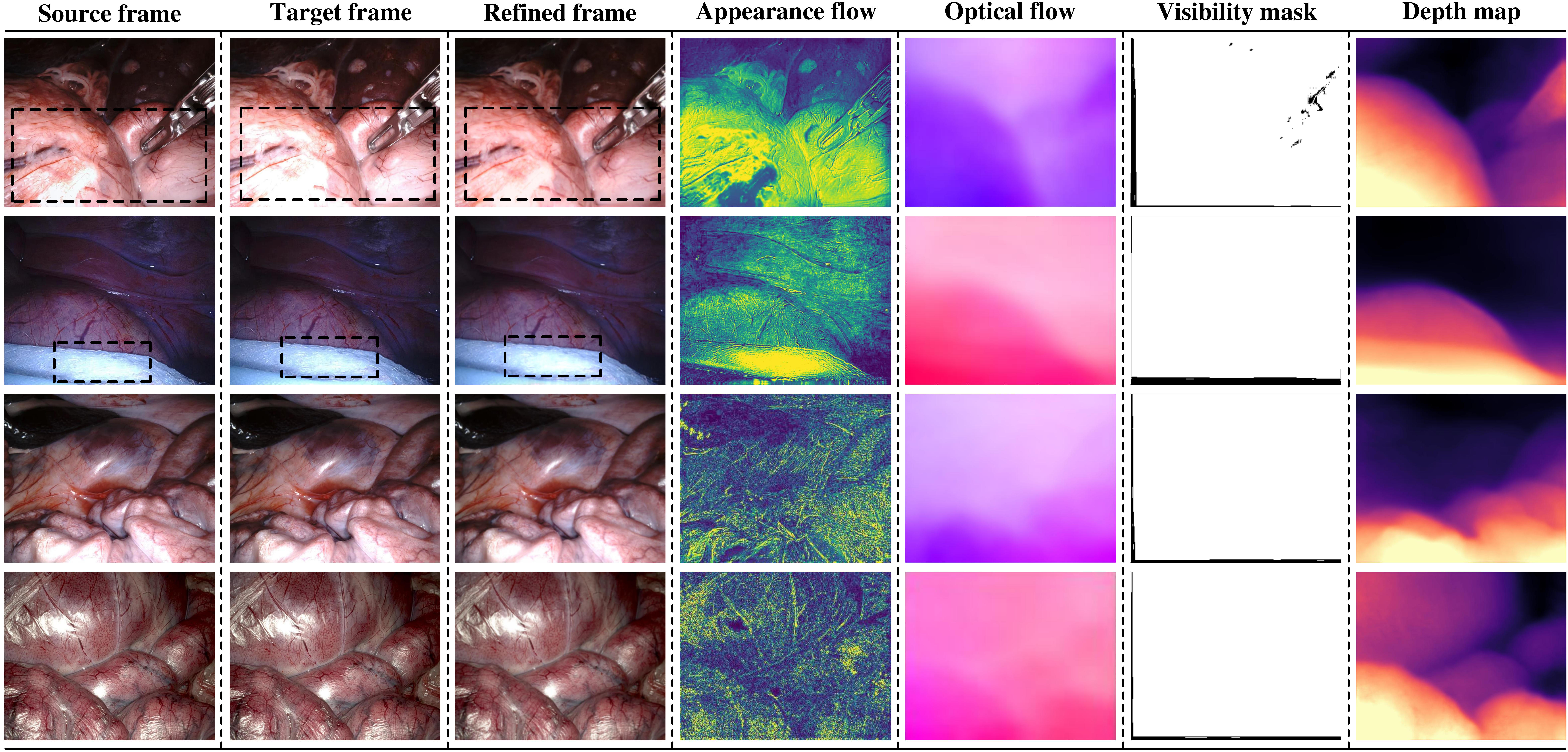}% 1\linewidth
		\caption{Visualization of the refined frames, appearance flows, optical flows, visibility masks and depth maps. Initially, the source and target frames have different brightness, particularly in the regions highlighted by black boxes. With the brightness calibration procedure, the refined frames appear to have the similar brightness as the source frames, which facilitates monocular depth and ego-motion self-supervised learning. The appearance flows have high values in regions close to the camera and  directly illuminated by the light source. Meanwhile, the appearance flows also maintain minor or zero values in the brightness constancy regions.}
		\label{Fig3}
	\end{figure*}

	In Fig.~\ref{Fig3}, we present some qualitative results of the refined frames, appearance flows, optical flows, visibility masks and depth maps.
	
	\section{Experiment}
	
		\subsection{Datasets}
		\begin{itemize}
			\item SCARED~\citep{allan2021stereo}. The SCARED dataset is collected from fresh porcine cadaver abdominal anatomy and contains 35 endoscopic videos with the point cloud and ego-motion ground truth.
			\item EndoSLAM~\citep{bengisu2020quantitative}. The EndoSLAM dataset is acquired from ex vivo porcine gastrointestinal tract organs and contains challenging ex vivo parts with pose ground truths and synthetic parts with depth and pose ground truths.
			\item SERV-CT~\citep{edwards2020serv}. SERV-CT contains16 stereo pairs collected from ex vivo porcine torso cadavers, along with the depth and disparity ground truths.
			\item Hamlyn\footnote{\url{http://hamlyn.doc.ic.ac.uk/vision/}} involves phantom heart model videos with the point cloud ground truth and in vivo endoscopic videos taken from various surgical procedures.
	\end{itemize}
	
	For depth estimation, we perform extensive experiments on the SCARED dataset and then apply models trained on the SCARED to the SERV-CT dataset and Hamlyn dataset to validate the generalization ability. In addition, we show some qualitative depth results on the EndoSLAM dataset in Fig.~\ref{Fig4}. We refer to the Eigen-Zhou evaluation protocol~\citep{eigen2014depth,zhou2017unsupervised} established on the KITTI benchmark and split the SCARED dataset into 15351, 1705, and 551 frames for the training, validation and test sets, respectively. We conduct ego-motion estimation on the SCARED dataset and the EndoSLAM dataset. We select two consecutive trajectories of 410 and 833 frames to evaluate performance on the SCARED dataset. In terms of the EndoSLAM dataset, we conduct 5-fold cross validation on the challenging LowCam of the ex vivo part, and each of the 5 models is tested on trajectories from one organ and trained on the rest. Besides, we demonstrate quantitative point cloud results and qualitative surface reconstructions on the SCARED dataset and Hamlyn dataset in Table~\ref{table9}, Table~\ref{table10} and Fig.~\ref{Fig8}.
	
	\subsection{Experimental Settings}
	
	\noindent \textbf{Training.} Our framework is implemented in the PyTorch library~\citep{paszke2017automatic} and trained on a single Nvidia TITAN RTX. We use the Adam optimizer~\citep{kingma2014adam}, where ${\beta _1} = 0.9,{\beta _2} = 0.99$ and the batch size is 12. We can achieve our results with the following weights set: $\alpha  = 0.85$, $\kappa = 1$, $\lambda_1 = 0.01$, $\lambda_2 = 0.01$, $\lambda_3= 0.0001$. As with existing works~\citep{godard2019digging,bian2019depth,spencer2020defeat,bengisu2020quantitative}, we employ the ResNet-18 encoder pretrained on ImageNet~\citep{deng2009imagenet}, which has been shown to reduce training time and improve estimation accuracy. For all subnetworks, the resolution of the input is $320 \times 256$ pixels. The entire training schedule contains two stages. First, we only train OFNet in a self-supervised manner via data fidelity loss (without appearance flow term) and edge-aware smoothness loss. After 20 epochs, we freeze OFNet and train DepthNet, PoseNet and AFNet simultaneously for another 20 epochs. In both stages, the initial learning rate is set to 1e-4 and multiplied by a scale factor of 0.1 after 10 epochs.
	More training schedules and corresponding impacts on depth estimation accuracy can be found in Table~\ref{table7}. ~\\
		\begin{table}[htb]
		\caption{The error and accuracy metrics for depth evaluation. $d$ and ${{d^ * }}$ respectively denotes the predicted depth value and the corresponding ground truth value, $\bm{{\rm{D}}}$ represents a set of predicted depth values.}
		\begin{center}
			%\smallskip
			\renewcommand{\arraystretch}{1.2}
			\resizebox{0.75\columnwidth}{!}{\begin{tabular}{c c}
					\hline
					Metric & Definition \\ 
					\hline
					\hline
					\cellcolor {orange!20} Abs Rel & $\frac{1}{{\left| \bm{{\rm{D}}} \right|}}\sum\limits_{d \in \bm{{\rm{D}}}} {{{\left| {{d^ * } - d} \right|} \mathord{\left/
								{\vphantom {{\left| {{d^ * } - d} \right|} {{d^ * }}}} \right.
								\kern-\nulldelimiterspace} {{d^ * }}}} $\\ 
					\cellcolor {orange!20} Sq Rel & $\frac{1}{{\left| \bm{{\rm{D}}} \right|}}\sum\limits_{d \in \bm{{\rm{D}}}} {{{{{\left| {{d^ * } - d} \right|}^2}} \mathord{\left/
								{\vphantom {{{{\left| {{d^ * } - d} \right|}^2}} {{d^ * }}}} \right.
								\kern-\nulldelimiterspace} {{d^ * }}}}$\\ 
					\cellcolor {orange!20} RMSE & $ \sqrt {\frac{1}{{\left| \bm{{\rm{D}}} \right|}}\sum\limits_{d \in \bm{{\rm{D}}}} {{{\left| {{d^ * } - d} \right|}^2}} }$\\
					
					\cellcolor {orange!20} RMSE log & $\sqrt {\frac{1}{{\left| \bm{{\rm{D}}} \right|}}\sum\limits_{d \in \bm{{\rm{D}}}} {{{\left| {\log {d^ * } - \log d} \right|}^2}} } $ \\
					\cellcolor {blue!20} $\delta$ & $\frac{1}{{\left| \bm{{\rm{D}}} \right|}}\left| {\left\{ {d \in \bm{{\rm{D}}}\left| {\max \left( {\frac{{{d^*}}}{d},\frac{d}{{{d^*}}} < 1.25} \right)} \right.} \right\}} \right. \times 100\% $ \\ 
					\hline
					
			\end{tabular}}
		\end{center}
		\label{table2}
	\end{table}  	
	
	\begin{table*}[htb!]
		\caption{Quantitative depth comparison on the SCARED dataset. The backbone of SfMLearner, Fang et al., DeFeat-Net, SC-SfMLearner, Monodepth2, Endo-SfM, baseline and ours are DispNet, VGG-16 BN, ResNet-18, ResNet-18, ResNet-18, ResNet-18, ResNet-18 and ResNet-18, respectively, which refers to the encoder of DepthNet. CIs is the abbreviation of confidence intervals. The paired p-values of our method and the compared methods are less than 0.05 on all metrics. The best results are in \textcolor{red}{red}. The second best are \uline{underlined}.}
		\begin{center}
			%\smallskip
			\renewcommand{\arraystretch}{1.3}
			\resizebox{2.0\columnwidth}{!}{\begin{tabular}{c c c c c c c c c c c}	
					\hline
					Method & \cellcolor {orange!20} Abs Rel $\downarrow$ & 95\% CIs & \cellcolor {orange!20} Sq Rel $\downarrow$ & 95\% CIs & \cellcolor {orange!20} RMSE $\downarrow$& 95\% CIs& \cellcolor {orange!20} RMSE log $\downarrow$ &95\% CIs & \cellcolor {blue!20} $\delta$ $\uparrow$ & 95\% CIs\\ 
					\hline						
					\hline
					SfMLearner (CVPR'17) & 0.079 & [0.076, 0.081] & 0.879  & [0.794, 0.964] & 6.896 & [6.513, 7.279] & 0.110 & [0.106, 0.115]& 0.947 & [0.942, 0.952]\\
					Fang et al. (WACV'20) & 0.078 &[0.076, 0.081] & 0.794& [0.737, 0.850] & 6.794& [6.482, 7.105] & 0.109 & [0.106, 0.113] & 0.946 & [0.941, 0.950]\\
					DeFeat-Net (CVPR'20) & 0.077 & [0.074, 0.079] & 0.792 & [0.731, 0.853] & 6.688 & [6.355, 7.021]& 0.108 & [0.104, 0.112] & 0.941 & [0.936, 0.946] \\
					SC-SfMLearner (NeurIPS'19) & 0.068 & [0.066, 0.071] & 0.645 & [0.589, 0.701] & 5.988 & [5.662, 6.314]& 0.097 &[0.093, 0.101] & \uline{0.957}&[0.953, 0.961]\\
					Monodepth2 (ICCV'19) & 0.071 & [0.068, 0.073] & \uline{0.590} & [0.554, 0.627]& \uline{5.606}& [5.404, 5.809] & 0.094 & [0.091, 0.097] & 0.953 & [0.948, 0.957]\\
					\hline
					Endo-SfM (MedIA'21)&  \uline{0.062} &[0.060, 0.065] & 0.606 & [0.551, 0.661] & 5.726 & [5.396, 6.056]& \uline{0.093} & [0.089, 0.097]& \uline{0.957} & [0.952, 0.961]\\
					\hline
					Baseline & 0.072 & [0.070, 0.074] & 0.614 & [0.579, 0.650] & 5.891 & [5.665, 6.117]& 0.099&[0.096, 0.102] & 0.949 & [0.944, 0.954]\\
					Ours & \textcolor{red}{0.059}& [0.057, 0.061]& \textcolor{red}{0.435} & [0.406, 0.464]& \textcolor{red}{4.925}& [4.729, 5.122]& \textcolor{red}{0.082} & [0.079, 0.084]& \textcolor{red}{0.974}& [0.971, 0.977]\\
					\hline
			\end{tabular}}
		\end{center}
		\label{table3}
	\end{table*}
	\begin{table*}[htb!]
		\caption{ SERV-CT depth results. All methods are self-supervised monocular trained on the SCARED dataset. The paired p-values of our method and the compared methods are less than 0.05 on all metrics. 
		}
		\begin{center}
			%\smallskip
			\renewcommand{\arraystretch}{1.3}
			\resizebox{2.0\columnwidth}{!}{\begin{tabular}{c c c c c c c c c c c}	
					\hline
					Method & \cellcolor {orange!20} Abs Rel $\downarrow$ & 95\% CIs & \cellcolor {orange!20} Sq Rel $\downarrow$ & 95\% CIs & \cellcolor {orange!20} RMSE $\downarrow$& 95\% CIs& \cellcolor {orange!20} RMSE log $\downarrow$ &95\% CIs & \cellcolor {blue!20} $\delta$ $\uparrow$ & 95\% CIs\\ 
					\hline						
					\hline
					SfMLearner (CVPR'17) & 0.151 & [0.137, 0.165] & 3.917  & [3.155, 4.680] & 17.451 & [15.298, 19.604] & 0.191 & [0.176, 0.207]& 0.779 & [0.741, 0.817]\\
					Fang et al. (WACV'20) & 0.149 &[0.140, 0.158] & 3.099& [2.652, 3.546] & 15.564& [14.041, 17.088] & 0.188 & [0.176, 0.200] & 0.787 & [0.758, 0.816]\\
					DeFeat-Net (CVPR'20) &  \uline{0.114} & [0.105, 0.124] &  \uline{1.946} & [1.534, 2.358] & 12.588 & [10.888, 14.287]& 0.153 & [0.139, 0.166] &  \uline{0.873} & [0.843, 0.902] \\
					SC-SfMLearner (NeurIPS'19) & 0.117 & [0.105, 0.129] & 2.015 & [1.568, 2.462] &  \uline{12.415} & [10.664, 14.167]& 0.148 &[0.134, 0.163] & 0.852&[0.816, 0.888]\\
					Monodepth2 (ICCV'19)& 0.123 & [0.112, 0.134] & 2.205 & [1.727, 2.683]& 12.927& [11.296, 14.557] & 0.152 & [0.138, 0.165] & 0.856 & [0.824, 0.888]\\
					\hline
					Endo-SfM (MedIA'21) &  0.116 &[0.105, 0.127] & 2.014 & [1.610, 2.419] & 12.493 & [10.921, 14.066]&  \uline{0.143} & [0.130, 0.157]& 0.864 & [0.829, 0.899]\\
					\hline
					Baseline & 0.129 & [0.119, 0.138] & 2.213 & [1.814, 2.612] & 13.095 & [11.503, 14.688]& 0.163&[0.151, 0.174] & 0.827 & [0.796, 0.858]\\
					Ours & \textcolor{red}{0.102}& [0.091, 0.113]& \textcolor{red}{1.632} & [1.235, 2.029]& \textcolor{red}{11.092}& [9.432, 12.751]& \textcolor{red}{0.131} & [0.116, 0.145]& \textcolor{red}{0.898}& [0.868, 0.929]\\
					\hline
			\end{tabular}}
		\end{center}
		\label{table13}
	\end{table*}
	\begin{table*}[htb!]
		\caption{Hamlyn depth results.  Similar to the generalization ability study on SERV-CT dataset in Table~\ref {table13}, we use models trained on the SCARED dataset. The results are reported on phantom images. The paired p-values of our method and the compared methods are less than 0.05 on all metrics.}
		\begin{center}
			%\smallskip
			\renewcommand{\arraystretch}{1.3}
			\resizebox{2.0\columnwidth}{!}{\begin{tabular}{ c c c c c c c c c c c}
					\hline
					\hline
					Method & \cellcolor {orange!20} Abs Rel $\downarrow$ & 95\% CIs & \cellcolor {orange!20} Sq Rel $\downarrow$ & 95\% CIs & \cellcolor {orange!20} RMSE $\downarrow$& 95\% CIs& \cellcolor {orange!20} RMSE log $\downarrow$ &95\% CIs & \cellcolor {blue!20} $\delta$ $\uparrow$ & 95\% CIs\\ 
					\hline						
					\hline
					SfMLearner (CVPR'17)& 0.090 & [0.088, 0.093]& 1.359 & [1.304, 1.414]& 10.876 & [10.626, 11.127]& 0.147 &[0.144, 0.151]& 0.910 & [0.904, 0.915]\\
					Fang et al. (WACV'20)& 0.142 & [0.139, 0.145]& 2.090 & [2.023, 2.156]& 12.666 & [12.443, 12.888]& 0.189 &[0.185, 0.192] & 0.810 & [0.799, 0.821]\\
					
					DeFeat-Net (CVPR'20)& 0.092 & [0.090, 0.094]& 1.174 &[1.126, 1.222]& 9.902 & [9.623, 10.181]& 0.138 &[0.135, 0.142]& 0.922&[0.917, 0.927] \\
					
					SC-SfMLearner (NeurIPS'19)& \uline{0.081}&[0.079, 0.083] & \uline{0.975} &[0.931, 1.019]& 9.130 &[8.891, 9.368]& \uline{0.121} &[0.118, 0.124]& 0.944 &[0.940, 0.949]\\
					
					Monodepth2 (ICCV'19)& 0.100 & [0.097, 0.103]& 1.106 & [1.060, 1.153]& \uline{8.786} &[8.625, 8.947]& 0.126 & [0.123, 0.129]& 0.927 & [0.922, 0.932]\\
					\hline				
					Endo-SfM (MedIA'21)& 0.085 &[0.082, 0.087] & 0.977 &[0.934, 1.021]& 8.960 &[8.714, 9.207]& 0.122 &[0.119, 0.125]& \uline{0.948}&[0.943, 0.953] \\
					\hline
					Baseline & 0.118 &[0.114, 0.121] & 1.455 &[1.395, 1.514]& 10.590 &[10.412, 10.768]& 0.159 &[0.154, 0.163]& 0.883 &[0.870, 0.896]\\
					Ours & \textcolor{red}{0.079} & [0.076, 0.081]& \textcolor{red}{0.800} &[0.772, 0.828]& \textcolor{red}{8.154} &[7.978, 8.330]& \textcolor{red}{0.109} &[0.107, 0.112]& \textcolor{red}{0.959}&[0.956, 0.963] \\
					\hline
			\end{tabular}}
		\end{center}
		\label{table4}
	\end{table*}
	
	\noindent \textbf{Performance metrics.} Table~\ref{table2} lists the depth evaluation metrics used in our experiments. During the validation, we scale the predicted depth maps with median scaling introduced by SFMLearner~\citep{zhou2017unsupervised}, which can be expressed as
	\begin{equation}
		\resizebox{0.85\hsize}{!}{$
			{\bm{{\rm{D}}}_{scaled}}\! =\! \left( {{\bm{{\rm{D}}}_{pred}}\! *\! \left( {{{median\left( {{\bm{{\rm{D}}}_{gt}}} \right)} \mathord{\left/
							{\vphantom {{median\left( {{\bm{{\rm{D}}}_{gt}}} \right)} {median\left( {{\bm{{\rm{D}}}_{pred}}} \right)}}} \right.
							\kern-\nulldelimiterspace} {median\left( {{\bm{{\rm{D}}}_{pred}}} \right)}}} \right)} \right)$}.
	\end{equation}
	The scaled depth maps are capped at 150 mm on the SCARED dataset and Hamlyn dataset and 180 mm on the SERV-CT dataset. A range of 150 mm and 180 mm can cover almost all depth values.
	
	For ego-motion, we adopt \citet{zhou2017unsupervised}'s 5-frame pose evaluation and the metric of absolute trajectory error (ATE)~\citep{mur2015orb}.
	
	\begin{table*}[htb!]
		\caption{Ablation study on the explored paradigms. AM: appearance module; CM: correspondence module; ${L_{rs}}$: residual-based smoothness loss; ${L_{ax}}$: auxiliary loss; ${L_{es}}$: edge-aware smoothness loss; ${\bm{{\rm{V}}}}\left( \bm{{\rm{p}}} \right)$: visibility mask. }
		\begin{center}
			%\smallskip
			\renewcommand{\arraystretch}{1.3}
			\resizebox{2.0\columnwidth}{!}{\begin{tabular}{c c c c c c c c c c c c c c c c}
					\hline
					ID & &AM&CM& ${L_{rs}}$ & ${L_{ax}}$ & \cellcolor {orange!20} Abs Rel $\downarrow$ & 95\% CIs &  \cellcolor {orange!20} Sq Rel $\downarrow$ & 95\% CIs & \cellcolor {orange!20} RMSE $\downarrow$ & 95\% CIs & \cellcolor {orange!20} RMSE log $\downarrow$ & 95\% CIs& \cellcolor {blue!20}$\delta$ $\uparrow$ & 95\% CIs\\ 
					\hline
					\hline				
					1 &&&&& & 0.072 & [0.070, 0.074]& 0.614 &[0.579, 0.650]& 5.891 &[5.665, 6.117]& 0.099 &[0.096, 0.102]& 0.949& [0.944, 0.954]\\
					2 && \cmark & & && 0.073 &[0.071, 0.075]& 0.626 &[0.589, 0.664]& 5.987 &[5.752, 6.223]& 0.099 &[0.096, 0.102]& 0.950&[0.945, 0.955]\\	
					3 &&\cmark&\cmark&&& 0.061 &[0.059, 0.063] & 0.472 &[0.442, 0.502]& 5.127 &[4.916, 5.338]& 0.085 &[0.082, 0.088]& 0.967&[0.964, 0.971]\\
					4 &&\cmark&\cmark&\cmark&& 0.060 &[0.058, 0.062]& 0.458 &[0.427, 0.490]& 5.016 &[4.800, 5.233]& 0.083 &[0.080, 0.086]& 0.969&[0.966, 0.972]\\
					5 &&\cmark&\cmark&&\cmark& 0.060 &[0.059, 0.062]& 0.457 &[0.427, 0.488]& 5.056 &[4.849, 5.263]& 0.084 &[0.081, 0.087]& 0.971&[0.968, 0.974] \\
					6 &&\cmark&\cmark&\cmark&\cmark&  \textcolor{red}{0.059} &[0.057, 0.061]& \textcolor{red}{0.435} &[0.406, 0.464]& \textcolor{red}{4.925} &[4.729, 5.122]& \textcolor{red}{0.082} & [0.079, 0.084]& \textcolor{red}{0.974}& [0.971, 0.977]\\
					\hline						
					7 &w/o ${L_{es}}$ &\cmark&\cmark&\cmark&\cmark& 0.061 & [0.059, 0.063]& 0.458 &[0.429, 0.488]& 5.067 &[4.869, 5.265]& 0.084 &[0.082, 0.087]& 0.968& [0.964, 0.971]\\
					8 &w/o ${\bm{{\rm{V}}}}\left( \bm{{\rm{p}}} \right)$ &\cmark&\cmark&\cmark&\cmark&  0.061 & [0.059, 0.063]& 0.463 &[0.432, 0.494]& 5.095 &[4.890, 5.301]& 0.084 &[0.081, 0.087]& 0.970& [0.967, 0.974]\\
					\hline				
			\end{tabular}}
		\end{center}
		\label{table5}
	\end{table*}
	
	\begin{table*}[htb!]
		\caption{Ablation study on the developed variants. A(Y): affine brightness transformer of D3VO~\citep{yang2020d3vo}; A(O): affine brightness transformer of Endo-SfM~\citep{bengisu2020quantitative}. Notations of AM, CM, ${L_{rs}}$ and ${L_{ax}}$ are in Table~\ref{table5}. }
		\begin{center}
			%\smallskip
			\renewcommand{\arraystretch}{1.3}
			\resizebox{2.0\columnwidth}{!}{\begin{tabular}{c c c c c c c c c c c c c c c c c}
					\hline
					ID & A(Y) & A(O) &AM & CM & ${L_{rs}}$ & ${L_{ax}}$ & \cellcolor {orange!20} Abs Rel $\downarrow$  & 95\% CIs & \cellcolor {orange!20} Sq Rel $\downarrow$ & 95\% CIs & \cellcolor {orange!20} RMSE $\downarrow$ & 95\% CIs & \cellcolor {orange!20} RMSE log $\downarrow$ & 95\% CIs & \cellcolor {blue!20} $\delta$ $\uparrow$ & 95\% CIs \\ 
					\hline
					\hline
					1  &&&&&&&  0.072 & [0.070, 0.074]& 0.614 &[0.579, 0.650]& 5.891 &[5.665, 6.117]& 0.099 &[0.096, 0.102]& 0.949& [0.944, 0.954]\\
					2 &\cmark&&&&&&  0.069  &[0.066, 0.071]& 0.602  &[0.558, 0.647]& 5.570  &[5.334, 5.807]& 0.092  &[0.089, 0.095]& 0.955 &[0.950, 0.960]\\	
					3 &&\cmark&&&&& 0.066 &[0.063, 0.068]& 0.554 &[0.517, 0.590]& 5.547 &[5.304, 5.790]& 0.091 &[0.088, 0.094]& 0.959&[0.954, 0.963]\\	  
					4 &\cmark&&&\cmark&&\cmark& 0.065 &[0.063, 0.067]& 0.538 &[0.504, 0.573]& 5.379 &[5.170, 5.587]& 0.089 &[0.086, 0.091]& 0.961&[0.958, 0.965]\\
					5 &&\cmark&&\cmark&&& 0.066 &[0.064, 0.068]& 0.554 &[0.516, 0.591]& 5.526 &[5.288, 5.764]& 0.091 &[0.088, 0.094]& 0.956&[0.952, 0.961]\\
					6 &&&\cmark&\cmark&\cmark&\cmark& \textcolor{red}{0.059} &[0.057, 0.061]& \textcolor{red}{0.435} &[0.406, 0.464]& \textcolor{red}{4.925} &[4.729, 5.122]& \textcolor{red}{0.082} & [0.079, 0.084]& \textcolor{red}{0.974}& [0.971, 0.977]\\					
					7 &\cmark&&\cmark&\cmark&\cmark&\cmark& 0.062& [0.060, 0.064]& 0.479&[0.449, 0.510] & 5.196& [4.983, 5.408]& 0.086 &[0.084, 0.089]& 0.967&[0.964, 0.971]\\
					8 &&\cmark&\cmark&\cmark&\cmark&\cmark& 0.063 &[0.061, 0.065]& 0.500 &[0.467, 0.532]& 5.268 &[5.048, 5.487]& 0.088 &[0.085, 0.091]& 0.963&[0.959, 0.967]\\
					\hline
			\end{tabular}}
		\end{center}
		\label{table6}
	\end{table*}

	\subsection{SCARED Depth}
	We evaluate the depth estimation accuracy of our framework against several typical self-supervised methods, including SfMLearner~\citep{zhou2017unsupervised}\footnote{\url{https://github.com/tinghuiz/SfMLearner}}, Fang et al.~\citep{fang2020towards}\footnote{\url{https://github.com/zenithfang/supervised_dispnet}}, DeFeat-Net~\citep{spencer2020defeat}\footnote{\url{https://github.com/jspenmar/DeFeat-Net}}, SC-SfMLearner~\citep{bian2019depth}\footnote{\url{https://github.com/JiawangBian/SC-SfMLearner-Release}}, Monodepth2~\citep{godard2019digging}\footnote{\url{https://github.com/nianticlabs/monodepth2}}, and Endo-SfM~\citep{bengisu2020quantitative}\footnote{\url{https://github.com/CapsuleEndoscope/EndoSLAM}}. 
	We reproduce these methods with the original implementation. Note that for SC-SfMLearner, we adopt an updated version released in the official GitHub, which has higher accuracy than the NeurIPS version. Table~\ref{table3} shows the quantitative results, and our framework exceeds all of the compared methods by a significant margin. Especially notable are results on the metric Sq Rel. According to Table~\ref{table2}, Sq Rel is sensitive to large depth errors. The presence of severe brightness fluctuations, induced by illumination variations, non-Lambertian reflection and interreflection, tends to constitute an extremely biased supervisory signal in the short- or median-range regions. Our framework is able to cope well with severe brightness fluctuations, and hence, much lower results on Sq Rel are acquired. In Fig.~\ref{Fig4}, we show a qualitative comparison.
	
	We believe that the inferior performance of these methods lies in the weakened supervisory signal on challenging endoscopic scenes. SfMLearner and Fang et al. selected L1 loss to assess the appearance difference, which is adversely affected by brightness variations. SC-SfMLearner, Monodepth2, DeFeat-Net and Endo-SfM used a weighted average of L1 loss and SSIM loss. SSIM is less susceptible to variations in the brightness pattern but instead is not invariant. In addition, DeFeat-Net leveraged deep features to enhance the supervisory signal when the brightness constancy assumption was violated. The overall scarcity and homogeneity of texture in endoscopic images, however, is harmful to visual representation learning. Endo-SfM is specifically devised for endoscopic videos and takes the brightness inconsistency problem into account. Nonetheless, the affine brightness transformer proposed in Endo-SfM can only adapt to limited violations, \textit{e.g.}, intensity shift.
	
	\begin{figure}[htb!]
		\centering
		\includegraphics[width=1.0\linewidth]{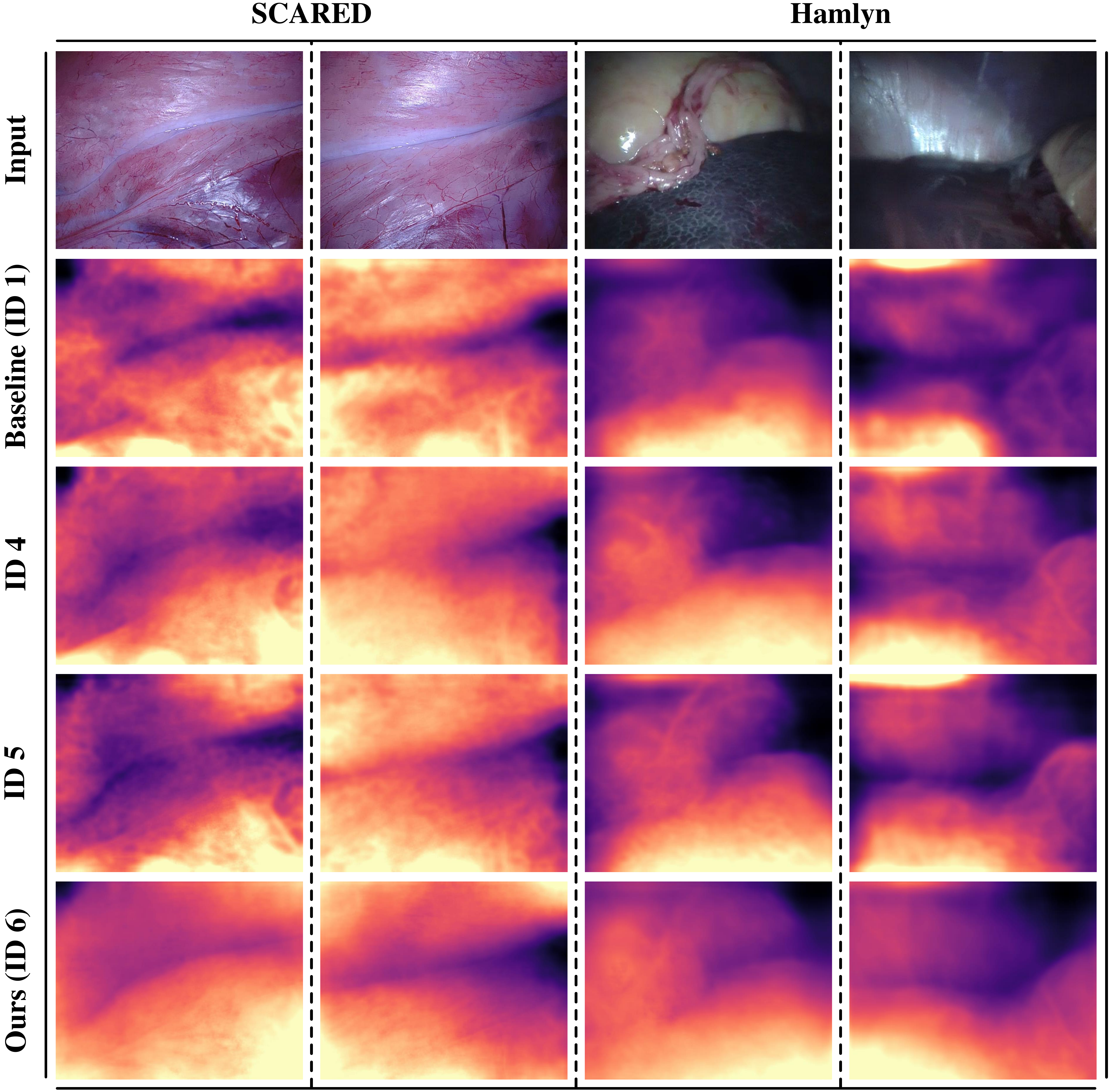}% 1\linewidth
		\caption{Qualitative results for baseline (ID 1), ID 4, ID 5 and ours (ID 6) in Table~\ref{table6}. Equipped with the appearance flow, our method can derive depth maps that are continuous and have fine-grained details, \textit{e.g.}, anatomical boundaries.}
		\label{Fig5}
	\end{figure}
	\begin{table*}[htb!]
		\caption{Ablation study on the different training schedules. For simplicity, we describe the combination of DepthNet and PoseNet, and DepthNet, PoseNet and AFNet as DPNet and DPANet, respectively. OFNet\dag: the OFNet is supervised via the data fidelity loss with appearance flow term involved. F (frozen): the weights of OFNet are frozen in the stage \uppercase\expandafter{\romannumeral2}. J (joint): the weights of OFNet are updated together with DPANet in one iteration.  A (alternating): the weights of OFNet are updated alternately with DPANet in one iteration. Concretely, in the first step, we freeze DPNAet's weights, and update OFNet. Reversely, we fix the weights of OFNet, and renew DPANet in the second step. In essence, the ``alternating" updating strategy is somewhat similar to the training mode of the Generative Adversarial Network~\citep{goodfellow2014generative}. }
		\begin{center}
			%\smallskip
			\renewcommand{\arraystretch}{1.3}
			\resizebox{2.0\columnwidth}{!}{\begin{tabular}{ccccccccccccccccc}
					\hline
					ID &Stage \uppercase\expandafter{\romannumeral1} &Stage \uppercase\expandafter{\romannumeral2} &Stage \uppercase\expandafter{\romannumeral3} &F & J &A & \cellcolor {orange!20} Abs Rel $\downarrow$ & 95\% CIs & \cellcolor {orange!20} Sq Rel $\downarrow$ & 95\% CIs & \cellcolor {orange!20} RMSE $\downarrow$ & 95\% CIs & \cellcolor {orange!20} RMSE log $\downarrow$ & 95\% CIs& \cellcolor {blue!20}$\delta$ $\uparrow$ & 95\% CIs\\  
					\hline
					\hline
					1 & DPNet & \xmark & \xmark &&&& 0.072 & [0.070, 0.074]& 0.614 &[0.579, 0.650]& 5.891 &[5.665, 6.117]& 0.099 &[0.096, 0.102]& 0.949& [0.944, 0.954]\\
					2 & DPANet, OFNet\dag & \xmark & \xmark &&\cmark&&  0.066 &[0.064, 0.069]& 0.535 &[0.502, 0.569]& 5.359 &[5.136, 5.581]& 0.091 &[0.087, 0.094]& 0.955&[0.950, 0.959]\\
					3 & DPANet, OFNet\dag &\xmark&\xmark&&&\cmark&  \textcolor{red}{0.059} &[0.057, 0.062]& 0.470 &[0.437, 0.504]& 5.062&[4.833, 5.291]& 0.083 &[0.080, 0.086]& 0.967&[0.963, 0.971]\\
					4 &OFNet\dag&DPANet&\xmark&\cmark&&&  0.065 &[0.063, 0.067]& 0.504 &[0.473, 0.535]& 5.366 &[5.161, 5.570]& 0.088 & [0.086, 0.091]& 0.967& [0.963, 0.970]\\
					5 &OFNet&DPANet&\xmark&\cmark&&&  \textcolor{red}{0.059} &[0.057, 0.061]& \textcolor{red}{0.435} &[0.406, 0.464]& \textcolor{red}{4.925} &[4.729, 5.122]& \textcolor{red}{0.082} & [0.079, 0.084]& \textcolor{red}{0.974}& [0.971, 0.977]\\
					6 &OFNet&DPANet&DPANet, OFNet\dag&\cmark&\cmark&&  0.064 &[0.061, 0.066]& 0.485 &[0.454, 0.515]& 5.146 &[4.948, 5.343]& 0.086&[0.083, 0.089] & 0.968&[0.964, 0.971] \\
					7 &OFNet&DPANet&DPANet, OFNet\dag&\cmark&&\cmark&  0.060 &[0.057, 0.062]& 0.460& [0.428, 0.493]& 4.988 &[4.776, 5.200]& 0.082&[0.079, 0.085] & 0.969&[0.966, 0.973]\\
					\hline
			\end{tabular}}
		\end{center}
		\label{table7}
	\end{table*}
	
	\begin{table*}[htb!]
		\caption{Quantitative ego-motion comparison on the EndoSLAM dataset. We adopt~\citet{zhou2017unsupervised}'s 5-frame pose evaluation, which averages the ATE over all 5-frame snippets. ``-'' means not applicable. The paired p-values of our method and Endo-SfM are 0.0890, 0.1707 and 0.3659 on trajectory-4 of Colon-IV, trajectory-3 of Stomach-I and trajectory-3 of Stomach-II, respectively. The p-values of our method and SC-SfMLearner are 0.4731, 0.3117 and 0.1256 on trajectory-5 of Colon-IV, trajectory-3 of Stomach-II and trajectory-2 of Stomach-III, respectively. Other than these cases, the p-values of our method and the compared methods are less than 0.05.}
		\begin{center}
			\smallskip
			\renewcommand{\arraystretch}{1.3}
			\resizebox{2.0\columnwidth}{!}{\begin{tabular}{c c c c c c c c c c c c}
					\hline
					&\multirow{2}{*}{Method} & \multicolumn{2}{c}{Trajectory-1 }&  \multicolumn{2}{c}{Trajectory-2 } & \multicolumn{2}{c}{Trajectory-3 }& \multicolumn{2}{c}{Trajectory-4 }& \multicolumn{2}{c}{Trajectory-5 } \\ 
					\cline{3-12} 
					&& ATE ($\times$ 1e-2)& 95\% CIs & ATE ($\times$ 1e-2) &  95\% CIs& ATE ($\times$ 1e-2) &  95\% CIs & ATE ($\times$ 1e-2) &  95\% CIs & ATE ($\times$ 1e-2) &  95\% CIs    \\ 
					\hline
					\hline
					\multirow{3}{*}{Colon-IV} & SC-SfMLearner & 0.2036 &[0.0569, 0.3504]& 0.1130&[0.0954, 0.1306] & \uline{0.0767}&[0.0688, 0.0845]& \uline{0.0966}&[0.0544, 0.1389]& \uline{0.1000}&[0.0897, 0.1102] \\
					& Endo-SfM & \uline{0.1996} &[0.0634, 0.3358]& \uline{0.1090}&[0.0915, 0.1265] & 0.0828&[0.0727, 0.0928]&  0.0981&[0.0480, 0.1482]& \textcolor{red}{0.0948}&[0.0838, 0.1059] \\
					& Ours &  \textcolor{red}{0.1297} &[0.0644, 0.1951]&  \textcolor{red}{0.0946}&[0.0813, 0.1080] &  \textcolor{red}{0.0674}&[0.0612, 0.0737]& \textcolor{red}{0.0802}&[0.0554, 0.1051]&  0.1002&[0.0908, 0.1095] \\
					\hline
					\multirow{3}{*}{Intestine} & SC-SfMLearner & 0.1678 &[0.1565, 0.1791]& 0.1478&[0.1339, 0.1618] & 0.1155&[0.1064, 0.1246]& 0.0957&[0.0692, 0.1221]& 0.6672&[0.6229, 0.7114] \\
					& Endo-SfM & \uline{0.1212} &[0.1123, 0.1301]& \uline{0.1205} &[0.1081, 0.1330] &  \textcolor{red}{0.0795}&[0.0719, 0.0871]&  \textcolor{red}{0.0663}&[0.0448, 0.0877]& \uline{0.5255} &[0.4876, 0.5633] \\
					& Ours &  \textcolor{red}{0.0894} &[0.0842, 0.0946]&  \textcolor{red}{0.0826}&[0.0753, 0.0898] & \uline{0.0930} &[0.0873, 0.0987]& \uline{0.0808} &[0.0643, 0.0974]&  \textcolor{red}{0.2502}&[0.2309, 0.2694] \\
					\hline
					\multirow{3}{*}{Stomach-I} & SC-SfMLearner & 0.1597 &[0.1396, 0.1798]& 0.1300&[0.1102, 0.1498] & 0.4473&[0.3436, 0.5509]& \uline{0.2176}&[0.1020, 0.3331]& -&- \\
					& Endo-SfM & \uline{0.1335} &[0.1137, 0.1533]& \uline{0.1230}&[0.1032, 0.1429] & \uline{0.4040}&[0.3211, 0.4870]&  \textcolor{red}{0.1893}&[0.0884, 0.2901]& -&- \\
					& Ours &  \textcolor{red}{0.1129} &[0.1002, 0.1256]&  \textcolor{red}{0.1098}&[0.0929, 0.1267] &  \textcolor{red}{0.3834}&[0.2885, 0.4782]& 0.2276&[0.1110, 0.3442]& -&- \\
					\hline
					\multirow{3}{*}{Stomach-II} & SC-SfMLearner & \uline{0.0989} &[0.0907, 0.1071]&  0.0580&[0.0243, 0.0917] & \textcolor{red}{0.1140} &[0.1017, 0.1262]& 0.1281&[0.1155, 0.1408]& -&- \\
					& Endo-SfM & 0.1047 &[0.0949, 0.1145]& \uline{0.0514}&[0.0225, 0.0803] & 0.1153&[0.1030, 0.1276]& \uline{0.1279} &[0.1156, 0.1402]& -&- \\
					& Ours &  \textcolor{red}{0.0917} &[0.0836, 0.0999]& \textcolor{red}{0.0457} &[0.0205, 0.0708] &  \uline{0.1148} &[0.1023, 0.1272]&  \textcolor{red}{0.1039}&[0.0947, 0.1132]& -&- \\
					\hline
					\multirow{3}{*}{Stomach-III} & SC-SfMLearner & 0.2022 &[0.1816, 0.2229]& \uline{0.1703} &[0.1547, 0.1860] & \uline{0.1878} &[0.1708, 0.2049]& 0.5120&[0.4657, 0.5583]& -&- \\
					& Endo-SfM & \uline{0.1993} &[0.1786, 0.2200]&  \textcolor{red}{0.1622}&[0.1480, 0.1764] & 0.1978&[0.1798, 0.2159]& \uline{0.5093} &[0.4645, 0.5541]& -&- \\
					& Ours &  \textcolor{red}{0.1558} &[0.1398, 0.1718]& 0.1750&[0.1581, 0.1918] &  \textcolor{red}{0.1497}&[0.1386, 0.1607]&  \textcolor{red}{0.4358}&[0.3940, 0.4777]& -&- \\
					\hline	
			\end{tabular}}
		\end{center}
		\label{table11}
	\end{table*}
	
	\begin{table}[htb!]
		\caption{Quantitative ego-motion comparison on the SCARED dataset. The ATE is averaged over all 5-frame snippets. The paired p-values of our method and the compared methods are less than 0.05. }
		\begin{center}
			%\smallskip
			\renewcommand{\arraystretch}{1.3}
			\resizebox{0.98\columnwidth}{!}{\begin{tabular}{c c c c c c }
					\hline
					\multirow{2}{*}{Method} & \multirow{2}{*}{Backbone} & \multicolumn{2}{c}{Trajectory-1 }&  \multicolumn{2}{c}{Trajectory-2 } \\ 
					\cline{3-6} 
					&& ATE& 95\% CIs & ATE & 95\% CIs    \\ 
					\hline
					\hline
					DeFeat-Net& ResNet-18 & 0.1765 &[0.1658, 0.1872]& 0.0995&[0.0953, 0.1037]\\
					SC-SfMLearner& ResNet-18 & 0.0767&[0.0715, 0.0818]& 0.0509& [0.0490, 0.0529]\\
					Monodepth2& ResNet-18 &
					0.0769&[0.0718, 0.0820]& 0.0554&[0.0532, 0.0576]\\
					\hline
					Endo-SfM& ResNet-18 & \uline{0.0759} & [0.0709, 0.0809]& \uline{0.0500} &[0.0480, 0.0519]\\			
					\hline
					Baseline  & ResNet-18 & 0.0814&[0.0762, 0.0865]& 0.0573 & [0.0551, 0.0595]\\		
					Ours  & ResNet-18 & \textcolor{red}{0.0742} &[0.0692, 0.0792]& \textcolor{red}{0.0478}& [0.0459, 0.0497] \\
					\hline
			\end{tabular}}
		\end{center}
		\label{table8}
	\end{table}
	
	\begin{table}[htb!]
		\caption{Quantitative average performance of ego-motion on each organ in the EndoSLAM dataset. The paired p-values of our method and the compared methods are less than 0.05. }
		\begin{center}
			%\smallskip
			\renewcommand{\arraystretch}{1.3}
			\resizebox{0.75\columnwidth}{!}{\begin{tabular}{c c c c }
					\hline
					&\multirow{2}{*}{Method} & \multicolumn{2}{c}{Trajectories} \\ 
					\cline{3-4} 
					&& ATE ($\times$ 1e-2)& 95\% CIs    \\ 
					\hline
					\hline
					\multirow{3}{*}{Colon-IV} & SC-SfMLearner & 0.1104 &[0.0882, 0.1325]\\
					& Endo-SfM & \uline{0.1096} &[0.0884, 0.1307]\\
					& Ours & \textcolor{red}{0.0917} &[0.0811, 0.1023]\\
					\hline
					\multirow{3}{*}{Intestine} & SC-SfMLearner & 0.2217 &[0.2105, 0.2328]\\
					& Endo-SfM & \uline{0.1690} &[0.1597, 0.1783] \\
					& Ours & \textcolor{red}{0.1133} &[0.1082, 0.1184]\\
					\hline
					\multirow{3}{*}{Stomach-I} & SC-SfMLearner & 0.1982 &[0.1746, 0.2217]\\
					& Endo-SfM & \uline{0.1790} &[0.1586, 0.1994] \\
					& Ours & \textcolor{red}{0.1662} &[0.1449, 0.1874]\\
					\hline
					\multirow{3}{*}{Stomach-II} & SC-SfMLearner & \uline{0.1137} &[0.1071, 0.1202]\\
					& Endo-SfM & 0.1158 &[0.1091, 0.1225] \\
					& Ours & \textcolor{red}{0.1035} &[0.0976, 0.1094]\\
					\hline
					\multirow{3}{*}{Stomach-III} & SC-SfMLearner & 0.2697 &[0.2549, 0.2844]\\
					& Endo-SfM & \uline{0.2690} &[0.2545, 0.2835]\\
					& Ours & \textcolor{red}{0.2301} &[0.2172, 0.2431]\\
					\hline	
			\end{tabular}}
		\end{center}
		\label{table12}
	\end{table}
	
	\begin{table}[!htb]
		\caption{Quantitative point cloud comparison on the SCARED dataset. Accuracy is a standard metric for point cloud evaluation~\citep{aanaes2016large}, and the results on other metrics are reported for reference. The paired p-values of our method and the compared methods are less than 0.05 on all metrics.}
		\begin{center}
			%\smallskip
			\renewcommand{\arraystretch}{1.3}
			\resizebox{1.0\columnwidth}{!}{\begin{tabular}{c c c c c c c}
					\hline
					Method & \cellcolor {orange!20} Accuracy $\downarrow$ &95\% CIs & \cellcolor {orange!20} Sq Rel $\downarrow$ &95\% CIs & \cellcolor {orange!20} RMSE log $\downarrow$ &95\% CIs\\ 
					\hline
					\hline
					SfMLearner& 4.446 &[4.192, 4.701] & 0.480 &[0.431, 0.528]& 0.120 &[0.116, 0.124]  \\
					Fang et al.& 4.384 &[4.177, 4.591]&0.431 &[0.399, 0.463]& 0.119&[0.116, 0.122] \\
					
					DeFeat-Net& 4.302 &[4.083, 4.520] & 0.424 &[0.390, 0.457]& 0.118 &[0.114, 0.122]\\
					
					SC-SfMLearner& 3.864 &[3.649, 4.079]& 0.351&[0.319, 0.382] & 0.108 &[0.105, 0.112] \\
					Monodepth2& \uline{3.614}&[3.482, 3.747] & \uline{0.319} &[0.299, 0.338]& \uline{0.105}&[0.102, 0.108] \\
					\hline
					Endo-SfM& 3.702  &[3.483, 3.920]& 0.331 &[0.299, 0.362]& \uline{0.105}&[0.101, 0.108] \\
					\hline
					Baseline  & 3.808&[3.659, 3.957]& 0.333&[0.314, 0.353] & 0.109&[0.106, 0.112] \\
					Ours  & \textcolor{red}{3.174} &[3.046, 3.302]& \textcolor{red}{0.234} &[0.218, 0.249]& \textcolor{red}{0.094} &[0.091, 0.096]\\
					\hline
			\end{tabular}}
		\end{center}
		\label{table9}
	\end{table}
	\begin{table}[!htb]
		\caption{Hamlyn point cloud results. All models trained by the SCARED are evaluated without any fine-tuning. The paired p-values of our method and the compared methods are less than 0.05 on all metrics.}
		\begin{center}
			%\smallskip
			\renewcommand{\arraystretch}{1.3}
			\resizebox{1.0\columnwidth}{!}{\begin{tabular}{c c  c c c c c}
					\hline
					Method & \cellcolor {orange!20} Accuracy $\downarrow$ &95\% CIs & \cellcolor {orange!20} Sq Rel $\downarrow$ &95\% CIs & \cellcolor {orange!20} RMSE log $\downarrow$ &95\% CIs\\ 
					\hline
					\hline
					SfMLearner& 6.968 &[6.808, 7.128]& 0.729 &[0.701, 0.757]& 0.160&[0.157, 0.164]    \\
					Fang et al.& 7.980 &[7.836, 8.123] &1.048 &[1.016, 1.079]& 0.198&[0.195, 0.202] \\
					
					DeFeat-Net& 6.303 &[6.126, 6.481] & 0.615 &[0.589, 0.641]& 0.152&[0.149, 0.155] \\
					
					SC-SfMLearner& 5.829 &[5.677, 5.981]& 0.511 &[0.489, 0.533]& 0.134&[0.131, 0.137] \\
					
					Monodepth2& \uline{5.472}&[5.372, 5.571]& 0.527 &[0.505, 0.549]& \uline{0.132}&[0.129, 0.135] \\
					\hline
					Endo-SfM& 5.689 &[5.532, 5.846]& \uline{0.500}&[0.478, 0.522] & 0.134 &[0.131, 0.136]\\
					\hline
					Baseline  & 6.655 &[6.542, 6.768]& 0.710 &[0.684, 0.736]& 0.168&[0.164, 0.171] \\
					Ours & \textcolor{red}{5.184} &[5.066, 5.302]& \textcolor{red}{0.412} &[0.398, 0.426]& \textcolor{red}{0.122} &[0.120, 0.123] \\
					\hline
			\end{tabular}}
		\end{center}
		\label{table10}
	\end{table}

	\subsection{SERV-CT and Hamlyn Depth}
\begin{figure*}[!htb]
		% Requires \usepackage{graphicx}
		\centering
		\includegraphics[width=1.0\linewidth]{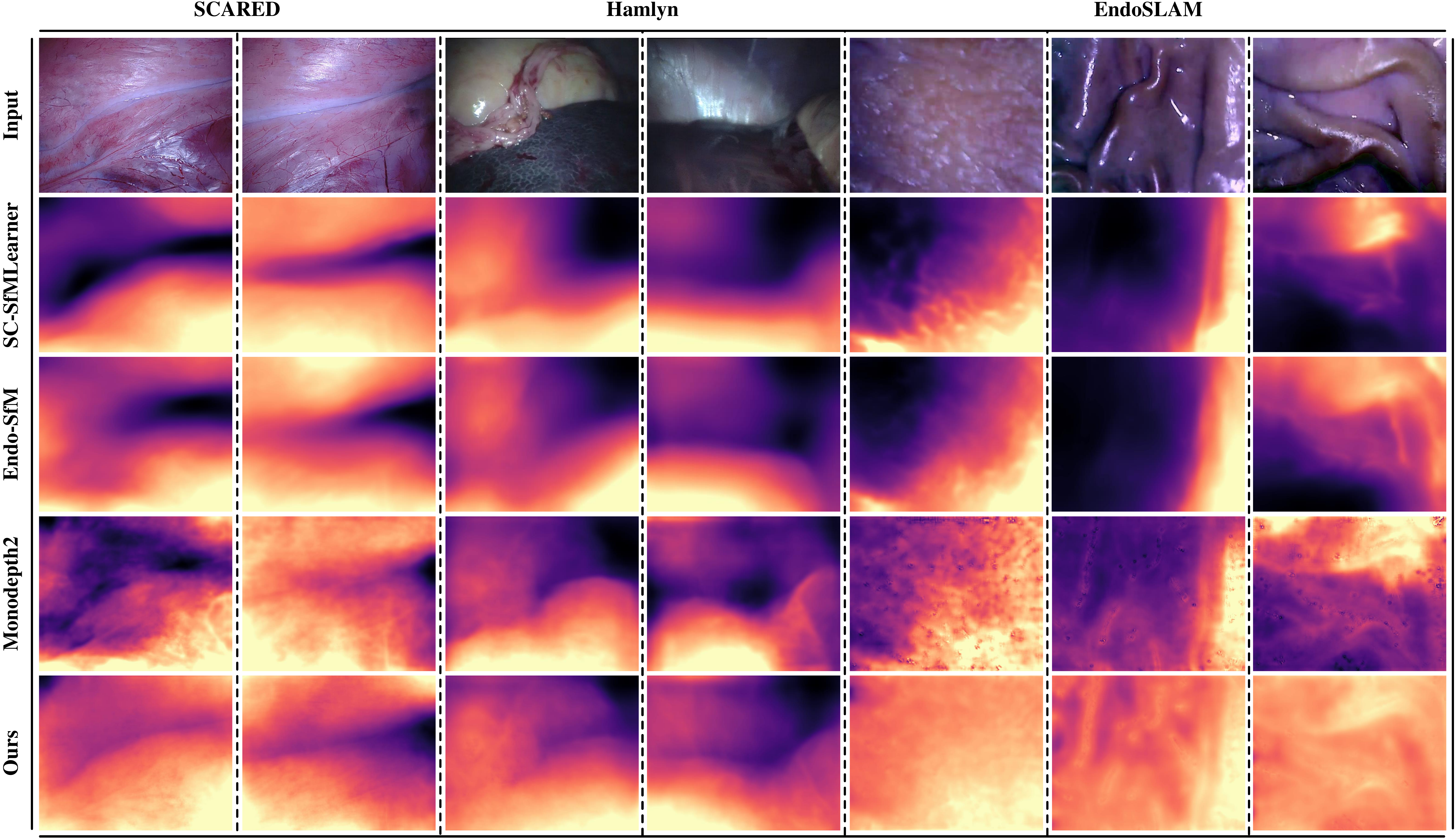}% 1\linewidth
		\caption{Qualitative depth comparison on the SCARED, Hamlyn and EndoSLAM datasets. Our method can generate more continuous depth maps, and is better at delineating anatomical structure than the compared methods, especially for the investigated cases from the challenging EndoSLAM dataset.}
		\label{Fig4}
	\end{figure*}

	We directly validate the models trained by the SCARED on the SERV-CT and Hamlyn datasets without any fine-tuning. We only adjust the resolution of test frames to $320 \times 256$ pixels and feed them to DepthNet. As presented in Table~\ref{table13} and Table~\ref{table4}, our framework achieves superior results than the other methods, revealing its strong generalization ability across different patients and cameras. A qualitative comparison can be seen in Fig.~\ref{Fig4}.
	
	\subsection{Ablation Study}
	To better understand how the diverse components contribute to the overall performance, we carry out a series of ablation studies at a more granular level, divided into explored paradigms, developed variants and training schedules. The results are reported with the scaled depth maps capped at 150 mm.~\\
	
	\noindent \textbf{Explored paradigm.} (Table~\ref{table5}) The baseline model (ID 1), with none of our contributions, performs poorly. Then, by directly adding the appearance module, we observe worse results on almost all evaluation metrics (ID 2). ID 3 shows the addition of the correspondence module, which greatly improves the evaluation performance over the baseline model. We then append the ${L_{rs}}$ (ID 4) or the ${L_{ax}}$ (ID 5), again with a steady improvement on all the evaluation metrics. With all of the contributions combined, we obtain the best results (ID 6). IDs 2, 3, 4, 5, and 6 indicate that the appearance flow, along with appropriate regularization, is the key to improving the depth estimation accuracy. Additionally, we ablate ${L_{es}}$ and ${\bm{{\rm{V}}}}\left( \bm{{\rm{p}}} \right)$ to understand the importance of edge-aware smoothness loss and visibility mask (IDs 7, 8), respectively.~\\	
	
	\noindent \textbf{Developed variants.} (Table~\ref{table6}) Considering the impacts of configuration differences on accuracy, we develop several variants to demonstrate the real strength of appearance flow with controlled experiments. We apply the affine brightness transformers proposed by~\citet{yang2020d3vo} (ID 2) and~\citet{bengisu2020quantitative} (ID 3) to the baseline model. This is done with settings in the original papers, \textit{e.g}, a pseudo-Siamese network is adopted to predict affine transformation parameters in ~\citep{yang2020d3vo}). In addition, we replace the appearance module with the affine brightness transformers in our framework (IDs 4, 5). Regardless of the deployment method, there still remains a large performance gap between variants IDs 2, 3, 4, 5 and our approach (ID 6). In Fig.~\ref{Fig5}, we present representative qualitative results of IDs 1, 4, 5, and 6.
	
	To a certain extent, brightness calibration is an analogous task to optical flow estimation or image registration, where large displacements between adjacent frames are one of the major difficulties. To address the thorny large displacements, recent deep learning-based methods~\citep{hur2019iterative,teed2020raft,shao2021multi,zhu2021test} share a common structure: from a coarse initial estimate, successive modules or subnetworks gradually refine the preceding estimates across pyramid levels or through cascaded subnetworks. Intuitively, we may handle severe brightness fluctuations in a similar manner. We assume that the brightness condition of the target frame and its counterpart is prealigned with the affine transformer of~\citep{yang2020d3vo} or~\citep{bengisu2020quantitative} so that the only source of brightness misalignment is nonlinear. Thereafter, we use the appearance module to handle the remaining part. As shown in Table~\ref{table6}, the cascaded strategy instead results in a decline in the depth estimation accuracy (IDs 7, 8). We contend that this is because in endoscopic environments, the appearance flow is already capable of compensating brightness variations for accurate depth estimation. In this case, inserting the affine calibration phase adds to the complexity of the task.~\\
	
	\begin{figure*}[!htb]
		% Requires \usepackage{graphicx}
		\centering
		\includegraphics[width=1.0\linewidth]{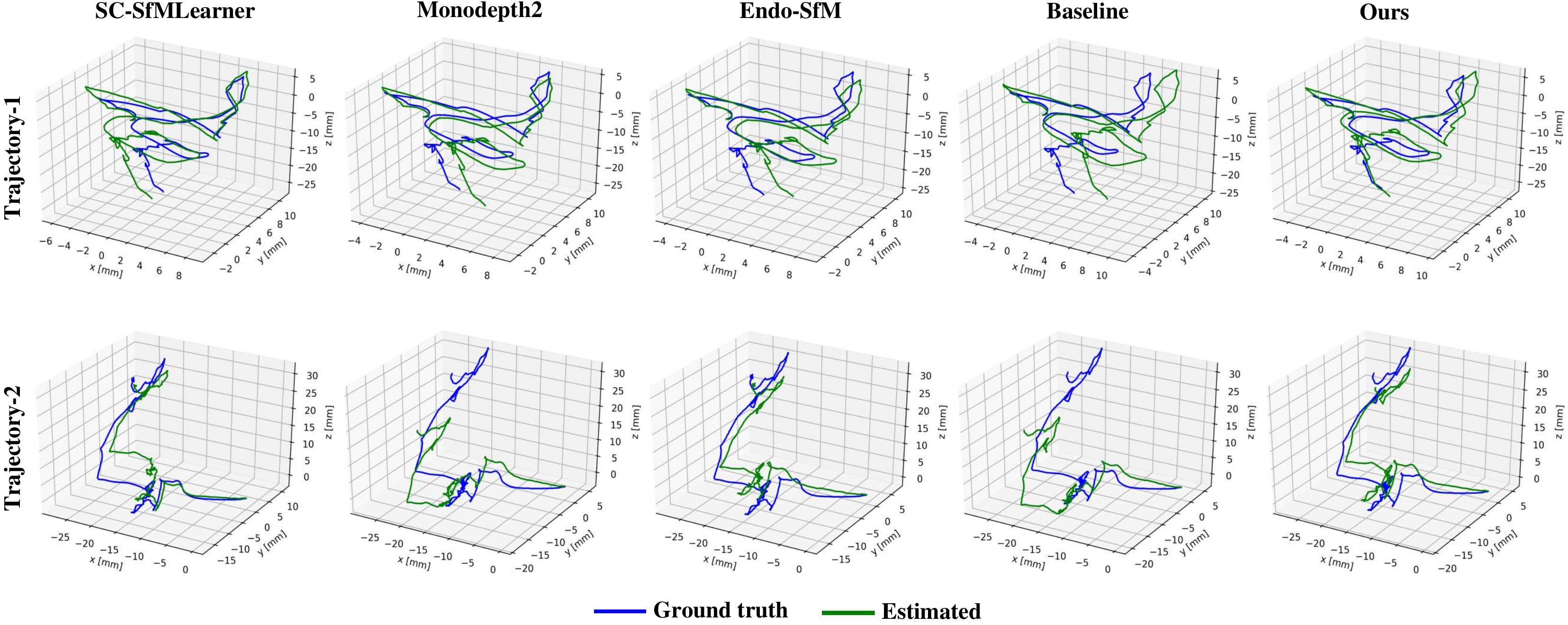}% 1\linewidth
		\caption{Qualitative comparison of the trajectory-1 (410 frames) and trajectory-2 (833 frames) from the SCARED dataset. Since the monocular deep visual odometry methods suffer from scale ambiguity, one calculated global scale over the whole sequence is used.}
		\label{Fig6}
	\end{figure*}
	
	\begin{figure*}[!htb]
		\centering
		\includegraphics[width=0.93\linewidth]{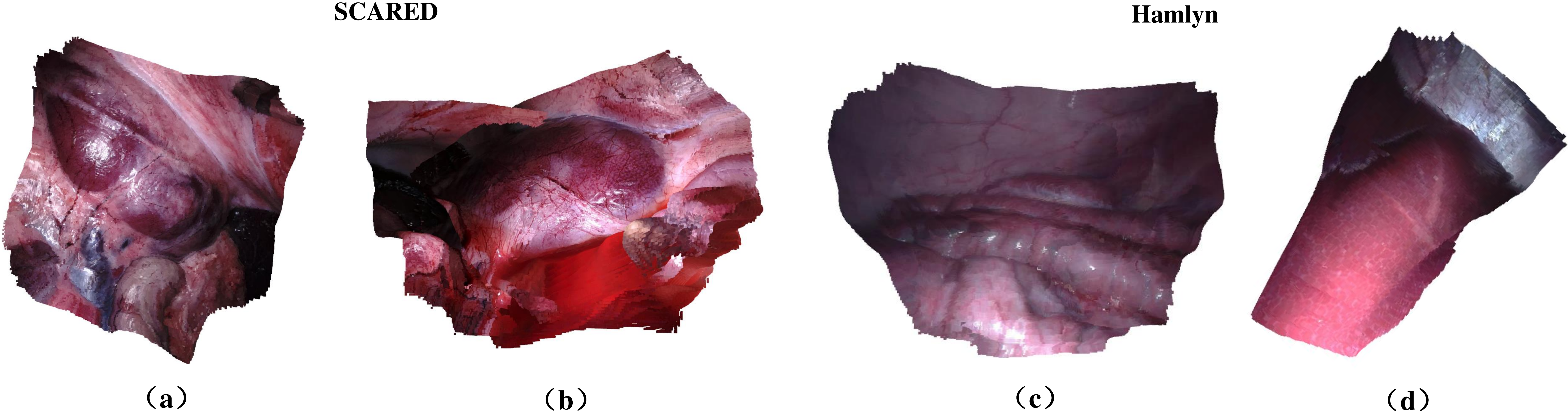}
		\caption{Our dense and accurate surface reconstructions on the SCARED dataset and Hamlyn dataset by using the model trained on the SCARED.}	
		\label{Fig8}
	\end{figure*}

	\noindent \textbf{Different training schedules.}
	(Table~\ref{table7}) Our unified framework comprises four subnetworks with intricate connections, exacerbating the nonconvergence potential of each subnetwork. Hence, it is critical to select a suitable training schedule. The explored training schedules can be categorized into two groups: end-to-end fashion (IDs 2, 3) and stagewise form (IDs 4, 5, 6, 7). Meanwhile, parameter updating strategies referred to as ``frozen", ``joint" and ``alternating" are also adopted when OFNet coexists with the other three subnetworks (DPANet). For ID 2 or ID 3, DPANet and OFNet are trained simultaneously for 20 epochs with the updating mode ``Joint" or ``Alternating". In terms of IDs 4, 5, we first train OFNet in a self-supervised manner. After 20 epochs, we freeze OFNet and train DPANet for another 20 epochs (``Frozen" mode). For ID 6 or ID 7, we use an additional training stage to fine-tune the overall network, and the learning process is 12 epochs. 
	
	We can clearly see that the training schedule is one of the key factors affecting the evaluation performance. DPANet relies on the output of OFNet, and there is no guarantee that the local gradient optimization could get the network to that optimal point with the updating mode ``Joint" (ID 2). It is worth mentioning that the alternative updating in one iteration can gradually approach the optimal point, allowing our framework to be trained in an end-to-end fashion (ID 3). When OFNet is trained separately, the appearance flow term included in the data fidelity loss makes it difficult for the OFNet to converge, which in turn affects the depth estimation accuracy in stage \uppercase\expandafter{\romannumeral2} (ID 4). A comparison between ID 5 and IDs 6, 7 indicates that the additional fine-tuning process seems invalid. Therefore, we choose a stagewise form up to stage \uppercase\expandafter{\romannumeral2}, which reduces computational cost and memory consumption at the same time (ID 5).
	
	 \subsection{SCARED and EndoSLAM Ego-Motion}
	Table~\ref{table8} shows a quantitative comparison on the SCARED dataset, and our method achieves lower ATE on both trajectories than several typical self-supervised methods. However, it seems that the accuracy improvement brought by our method in ego-motion is not as much as in depth, especially for results on trajectory 1. The reason behind this is that depth estimation is a task of pixel-to-pixel mapping, and observation noise in endoscopic videos, such as non-Lambertian reflection occurring in local regions, will greatly affect the depth map accuracy. For ego-motion estimation, PoseNet can utilize the feature vectors of all pixels when predicting each value in the $6$DOF pose, and thus ego-motion estimation has higher immunity against observation noise. Table~\ref{table11} and Table~\ref{table12} demonstrate the quantitative results of our method against the previous methods, Endo-SfM and SC-SfMLearner on the challenging EndoSLAM dataset. Table~\ref{table11} indicates that our method is capable of exceeding Endo-SfM and SC-SfMLearner for most of the cases. In Table~\ref{table12}, we further summarize the average performance on each organ. The superiority of our method in Table~\ref{table11} and Table~\ref{table12} verifies the effectiveness of the proposed appearance flow. In addition, we notice that even for folds that do not consist of the colon or small intestine in the training phase, our method achieves low ATE values in the colon or small intestine, which is an indicator of its excellent organ adaptability and generalization. 
	
	Fig.~\ref{Fig6} displays the trajectories predicted by SC-SfMLearner, Monodepth2, Endo-SfM, baseline and ours on trajectory-1 and trajectory-2 of the SCARED dataset. We can see that the trajectories of ours in general are better than those of the compared methods. It is interesting to note that Monodepth2 and the baseline exhibit a severe scale drift issue on trajectory-2, but ours succeeds even though these three frameworks share the same structure of PoseNet. We believe that this phenomenon arises from the appearance flow term introduced in the data fidelity loss, which is the main difference between our framework and the others. The appearance flow is capable of compensating the interframe brightness variations, and thus much more accurate predictions are achieved. From the quantitative and qualitative comparison of these three methods, the ego-motion accuracy and the scale consistency length may be positively correlated. Specifically, Monodepth2 and ours achieve higher accuracy and demonstrate less scale drift than the baseline on trajectory-1. Moreover, the accuracy of Monodepth2 and baseline is similar on trajectory-2, but our accuracy is much higher, and we obtain a globally scale-consistent trajectory.

	\subsection{SCARED and Hamlyn Point Cloud}
	The point clouds are converted from depth predictions with the camera intrinsic. To address the scale ambiguity, we adopt the same scaling strategy as in the depth evaluation. Table~\ref{table9} and Table~\ref{table10} summarize the quantitative results of point cloud reconstruction on the SCARED dataset and Hamlyn dataset, respectively. 
	In Fig.~\ref{Fig8}, we follow~\citep{recasens2021endo} and show high-quality surface reconstructions after fusing depth predictions in a truncated signed distance function (TSDF) representation~\citep{curless1996volumetric}. Although there is no ground truth for scene maps in the datasets of SCARED and Hamlyn, the accuracy of surface reconstructions can be qualitatively evaluated in Fig.~\ref{Fig8}, demonstrating that the reconstructions have few distortions and preserve prominent features, such as tissue contours. Specifically, the scene is first represented by a discrete voxel grid, and for each voxel, the weighted signed distance to the closest range surface is stored. The TSDF can be easily updated by using sequential averaging for each voxel and the depth predictions of each key frame. To obtain the final surfaces, we use a marching cubes algorithm~\citep{lorensen1987marching}. The implementation of surface reconstruction pipeline is available at \url{https://github.com/UZ-SLAMLab/Endo-Depth-and-Motion}.
	
	\section{Conclusion}
	In this work, we open up a new direction by introducing the concept of appearance flow to address severe interframe brightness fluctuations in endoscopic scenes. The appearance flow takes into account any variations in the brightness pattern and induces a generalized dynamic image constraint, allowing a relaxation of the brightness constancy assumption. We are confident that the introduced appearance flow can be applicable to more general scenarios, such as nighttime driving and adverse weather, or other tasks, such as stereo matching and multiview stereo. In addition, we present a novel unified self-supervised framework for robust depth and ego-motion estimation in minimally invasive surgery environments. In contrast to previously competing methods for self-supervised depth and ego-motion estimation, the framework has higher immunity against brightness variations. Finally, comprehensive experiments are conducted on multiple datasets, including SCARED, EndoSLAM, SERV-CT and Hamlyn. The experimental results illustrate the efficacy and generalization ability of our method.~\\

	\noindent \textbf{Limitations and future work.} There are limitations of our method that need to be taken into consideration in future work.
	First, our depth predictions in Fig.~\ref{Fig4} show slight artifacts in the oversaturated regions, which is caused by the losses of information due to oversaturation. The issue can be alleviated by increasing the weight of the smoothness loss term to enforce the depth propagation from meaningful regions to oversaturated regions. However, such propagation tends to yield oversmooth artifacts around the structural boundaries. In the future work, we will attempt to explore self-supervised monocular multiview depth estimation. It should be noted that the method in this work and the related methods, \textit{e.g.}, Endo-SfM~\citep{bengisu2020quantitative} and SC-SfMLearner~\citep{bian2019depth}, belong to the single-view depth estimation. We believe that utilizing the complementary information from multiple frames may overcome the artifacts brought by the oversaturated regions in the target frame. Second, our framework suffers from scale ambiguity when deriving predictions of depth maps and camera poses. In fact, the scale ambiguity is an intrinsic limitation that exists in the self-supervised monocular depth and ego-motion estimation algorithms. In the future work, we will investigate the possibilities of incorporating additional information, such as object size, to recover the absolute scale. Integrating the displacement or velocity of the endoscope that can be measured by tracking devices into the training phase might be a remedy. 
		
	\section{Declaration of Competing Interest}
		The authors declare that they have no known competing financial interests or personal relationships that could have appeared to influence the work reported in this paper.

	\section{Acknowledgement}
	This work was supported by the Key Research and Development Program of Shandong Province under Grant No. 2019JZZY011101, the National Natural Science Foundation of China under grants 61620106012 and 62076016, and the Shenzhen Science and Technology Program under grant No. KQTD2016112515134654, and the Key Research and Development Program of Zhejiang Province under Grant 2020C01109.

	\bibliographystyle{model2-names.bst}\biboptions{authoryear}
	\bibliography{refers}

\begin{thebibliography}{70}
\expandafter\ifx\csname natexlab\endcsname\relax\def\natexlab#1{#1}\fi
\providecommand{\url}[1]{\texttt{#1}}
\providecommand{\href}[2]{#2}
\providecommand{\path}[1]{#1}
\providecommand{\DOIprefix}{doi:}
\providecommand{\ArXivprefix}{arXiv:}
\providecommand{\URLprefix}{URL: }
\providecommand{\Pubmedprefix}{pmid:}
\providecommand{\doi}[1]{\href{http://dx.doi.org/#1}{\path{#1}}}
\providecommand{\Pubmed}[1]{\href{pmid:#1}{\path{#1}}}
\providecommand{\bibinfo}[2]{#2}
\ifx\xfnm\relax \def\xfnm[#1]{\unskip,\space#1}\fi
%Type = Article
\bibitem[{Aan{\ae}s et~al.(2016)Aan{\ae}s, Jensen, Vogiatzis, Tola and
  Dahl}]{aanaes2016large}
\bibinfo{author}{Aan{\ae}s, H.}, \bibinfo{author}{Jensen, R.R.},
  \bibinfo{author}{Vogiatzis, G.}, \bibinfo{author}{Tola, E.},
  \bibinfo{author}{Dahl, A.B.}, \bibinfo{year}{2016}.
\newblock \bibinfo{title}{Large-scale data for multiple-view stereopsis}.
\newblock \bibinfo{journal}{International Journal of Computer Vision}
  \bibinfo{volume}{120}, \bibinfo{pages}{153--168}.
%Type = Article
\bibitem[{Allan et~al.(2021)Allan, Mcleod, Wang, Rosenthal, Fu, Zeffiro, Xia,
  Zhanshi, Luo and Zhang}]{allan2021stereo}
\bibinfo{author}{Allan, M.}, \bibinfo{author}{Mcleod, J.},
  \bibinfo{author}{Wang, C.C.}, \bibinfo{author}{Rosenthal, J.C.},
  \bibinfo{author}{Fu, K.X.}, \bibinfo{author}{Zeffiro, T.},
  \bibinfo{author}{Xia, W.}, \bibinfo{author}{Zhanshi, Z.},
  \bibinfo{author}{Luo, H.}, \bibinfo{author}{Zhang, X.}, \bibinfo{year}{2021}.
\newblock \bibinfo{title}{Stereo correspondence and reconstruction of
  endoscopic data challenge}.
\newblock \bibinfo{journal}{arXiv preprint arXiv:2101.01133} .
%Type = Article
\bibitem[{Bernhardt et~al.(2017)Bernhardt, Nicolau, Soler and
  Doignon}]{bernhardt2017status}
\bibinfo{author}{Bernhardt, S.}, \bibinfo{author}{Nicolau, S.A.},
  \bibinfo{author}{Soler, L.}, \bibinfo{author}{Doignon, C.},
  \bibinfo{year}{2017}.
\newblock \bibinfo{title}{The status of augmented reality in laparoscopic
  surgery as of 2016}.
\newblock \bibinfo{journal}{Medical Image Analysis} \bibinfo{volume}{37},
  \bibinfo{pages}{66--90}.
%Type = Inproceedings
\bibitem[{Bian et~al.(2019)Bian, Li, Wang, Zhan, Shen, Cheng and
  Reid}]{bian2019depth}
\bibinfo{author}{Bian, J.W.}, \bibinfo{author}{Li, Z.}, \bibinfo{author}{Wang,
  N.}, \bibinfo{author}{Zhan, H.}, \bibinfo{author}{Shen, C.},
  \bibinfo{author}{Cheng, M.M.}, \bibinfo{author}{Reid, I.},
  \bibinfo{year}{2019}.
\newblock \bibinfo{title}{Unsupervised scale-consistent depth and ego-motion
  learning from monocular video}, in: \bibinfo{booktitle}{Thirty-third
  Conference on Neural Information Processing Systems}.
%Type = Article
\bibitem[{Cao et~al.(2017)Cao, Wu and Shen}]{cao2017estimating}
\bibinfo{author}{Cao, Y.}, \bibinfo{author}{Wu, Z.}, \bibinfo{author}{Shen,
  C.}, \bibinfo{year}{2017}.
\newblock \bibinfo{title}{Estimating depth from monocular images as
  classification using deep fully convolutional residual networks}.
\newblock \bibinfo{journal}{IEEE Transactions on Circuits and Systems for Video
  Technology} \bibinfo{volume}{28}, \bibinfo{pages}{3174--3182}.
%Type = Inproceedings
\bibitem[{Casser et~al.(2019)Casser, Pirk, Mahjourian and
  Angelova}]{casser2019depth}
\bibinfo{author}{Casser, V.}, \bibinfo{author}{Pirk, S.},
  \bibinfo{author}{Mahjourian, R.}, \bibinfo{author}{Angelova, A.},
  \bibinfo{year}{2019}.
\newblock \bibinfo{title}{Depth prediction without the sensors: Leveraging
  structure for unsupervised learning from monocular videos}, in:
  \bibinfo{booktitle}{Proceedings of the AAAI Conference on Artificial
  Intelligence}, pp. \bibinfo{pages}{8001--8008}.
%Type = Article
\bibitem[{Chand et~al.(2021)Chand, Birlo and Stoyanov}]{chand2021challenge}
\bibinfo{author}{Chand, M.}, \bibinfo{author}{Birlo, M.},
  \bibinfo{author}{Stoyanov, D.}, \bibinfo{year}{2021}.
\newblock \bibinfo{title}{The challenge of augmented reality in surgery}.
\newblock \bibinfo{journal}{Digital Surgery} , \bibinfo{pages}{121--135}.
%Type = Article
\bibitem[{Chen et~al.(2018)Chen, Tang, John, Wan and Zhang}]{chen2018slam}
\bibinfo{author}{Chen, L.}, \bibinfo{author}{Tang, W.}, \bibinfo{author}{John,
  N.W.}, \bibinfo{author}{Wan, T.R.}, \bibinfo{author}{Zhang, J.J.},
  \bibinfo{year}{2018}.
\newblock \bibinfo{title}{Slam-based dense surface reconstruction in monocular
  minimally invasive surgery and its application to augmented reality}.
\newblock \bibinfo{journal}{Computer methods and programs in biomedicine}
  \bibinfo{volume}{158}, \bibinfo{pages}{135--146}.
%Type = Article
\bibitem[{Chen et~al.(2019a)Chen, Bobrow, Athey, Mahmood and
  Durr}]{chen2019slam}
\bibinfo{author}{Chen, R.J.}, \bibinfo{author}{Bobrow, T.L.},
  \bibinfo{author}{Athey, T.}, \bibinfo{author}{Mahmood, F.},
  \bibinfo{author}{Durr, N.J.}, \bibinfo{year}{2019}a.
\newblock \bibinfo{title}{Slam endoscopy enhanced by adversarial depth
  prediction}.
\newblock \bibinfo{journal}{arXiv preprint arXiv:1907.00283} .
%Type = Article
\bibitem[{Chen et~al.(2016)Chen, Fu, Yang and Deng}]{chen2016single}
\bibinfo{author}{Chen, W.}, \bibinfo{author}{Fu, Z.}, \bibinfo{author}{Yang,
  D.}, \bibinfo{author}{Deng, J.}, \bibinfo{year}{2016}.
\newblock \bibinfo{title}{Single-image depth perception in the wild}.
\newblock \bibinfo{journal}{arXiv preprint arXiv:1604.03901} .
%Type = Inproceedings
\bibitem[{Chen et~al.(2019b)Chen, Schmid and Sminchisescu}]{chen2019self}
\bibinfo{author}{Chen, Y.}, \bibinfo{author}{Schmid, C.},
  \bibinfo{author}{Sminchisescu, C.}, \bibinfo{year}{2019}b.
\newblock \bibinfo{title}{Self-supervised learning with geometric constraints
  in monocular video: Connecting flow, depth, and camera}, in:
  \bibinfo{booktitle}{Proceedings of the IEEE/CVF International Conference on
  Computer Vision}, pp. \bibinfo{pages}{7063--7072}.
%Type = Inproceedings
\bibitem[{Curless and Levoy(1996)}]{curless1996volumetric}
\bibinfo{author}{Curless, B.}, \bibinfo{author}{Levoy, M.},
  \bibinfo{year}{1996}.
\newblock \bibinfo{title}{A volumetric method for building complex models from
  range images}, in: \bibinfo{booktitle}{Proceedings of the 23rd annual
  conference on Computer graphics and interactive techniques}, pp.
  \bibinfo{pages}{303--312}.
%Type = Inproceedings
\bibitem[{Deng et~al.(2009)Deng, Dong, Socher, Li, Li and
  Fei-Fei}]{deng2009imagenet}
\bibinfo{author}{Deng, J.}, \bibinfo{author}{Dong, W.},
  \bibinfo{author}{Socher, R.}, \bibinfo{author}{Li, L.J.},
  \bibinfo{author}{Li, K.}, \bibinfo{author}{Fei-Fei, L.},
  \bibinfo{year}{2009}.
\newblock \bibinfo{title}{Imagenet: A large-scale hierarchical image database},
  in: \bibinfo{booktitle}{2009 IEEE conference on computer vision and pattern
  recognition}, \bibinfo{organization}{Ieee}. pp. \bibinfo{pages}{248--255}.
%Type = Inproceedings
\bibitem[{Dosovitskiy et~al.(2015)Dosovitskiy, Fischer, Ilg, Hausser, Hazirbas,
  Golkov, Van Der~Smagt, Cremers and Brox}]{dosovitskiy2015flownet}
\bibinfo{author}{Dosovitskiy, A.}, \bibinfo{author}{Fischer, P.},
  \bibinfo{author}{Ilg, E.}, \bibinfo{author}{Hausser, P.},
  \bibinfo{author}{Hazirbas, C.}, \bibinfo{author}{Golkov, V.},
  \bibinfo{author}{Van Der~Smagt, P.}, \bibinfo{author}{Cremers, D.},
  \bibinfo{author}{Brox, T.}, \bibinfo{year}{2015}.
\newblock \bibinfo{title}{Flownet: Learning optical flow with convolutional
  networks}, in: \bibinfo{booktitle}{Proceedings of the IEEE international
  conference on computer vision}, pp. \bibinfo{pages}{2758--2766}.
%Type = Article
\bibitem[{Edwards et~al.(2021)Edwards, Psychogyios, Speidel, Maier-Hein and
  Stoyanov}]{edwards2020serv}
\bibinfo{author}{Edwards, P.E.}, \bibinfo{author}{Psychogyios, D.},
  \bibinfo{author}{Speidel, S.}, \bibinfo{author}{Maier-Hein, L.},
  \bibinfo{author}{Stoyanov, D.}, \bibinfo{year}{2021}.
\newblock \bibinfo{title}{Serv-ct: A disparity dataset from cone-beam cone-beam
  ct for validation of endoscopic 3d reconstruction}.
\newblock \bibinfo{journal}{Medical Image Analysis} , \bibinfo{pages}{102302}.
%Type = Inproceedings
\bibitem[{Eigen et~al.(2014)Eigen, Puhrsch and Fergus}]{eigen2014depth}
\bibinfo{author}{Eigen, D.}, \bibinfo{author}{Puhrsch, C.},
  \bibinfo{author}{Fergus, R.}, \bibinfo{year}{2014}.
\newblock \bibinfo{title}{Depth map prediction from a single image using a
  multi-scale deep network}, in: \bibinfo{booktitle}{Advances in Neural
  Information Processing Systems}, pp. \bibinfo{pages}{2366--2374}.
%Type = Inproceedings
\bibitem[{Fang et~al.(2020)Fang, Chen, Chen and Gool}]{fang2020towards}
\bibinfo{author}{Fang, Z.}, \bibinfo{author}{Chen, X.}, \bibinfo{author}{Chen,
  Y.}, \bibinfo{author}{Gool, L.V.}, \bibinfo{year}{2020}.
\newblock \bibinfo{title}{Towards good practice for cnn-based monocular depth
  estimation}, in: \bibinfo{booktitle}{Proceedings of the IEEE Winter
  Conference on Applications of Computer Vision}, pp.
  \bibinfo{pages}{1091--1100}.
%Type = Inproceedings
\bibitem[{Fu et~al.(2018)Fu, Gong, Wang, Batmanghelich and Tao}]{fu2018deep}
\bibinfo{author}{Fu, H.}, \bibinfo{author}{Gong, M.}, \bibinfo{author}{Wang,
  C.}, \bibinfo{author}{Batmanghelich, K.}, \bibinfo{author}{Tao, D.},
  \bibinfo{year}{2018}.
\newblock \bibinfo{title}{Deep ordinal regression network for monocular depth
  estimation}, in: \bibinfo{booktitle}{Proceedings of the IEEE Conference on
  Computer Vision and Pattern Recognition}, pp. \bibinfo{pages}{2002--2011}.
%Type = Inproceedings
\bibitem[{Geiger et~al.(2012)Geiger, Lenz and Urtasun}]{geiger2012we}
\bibinfo{author}{Geiger, A.}, \bibinfo{author}{Lenz, P.},
  \bibinfo{author}{Urtasun, R.}, \bibinfo{year}{2012}.
\newblock \bibinfo{title}{Are we ready for autonomous driving? the kitti vision
  benchmark suite}, in: \bibinfo{booktitle}{Proceedings of the IEEE Conference
  on Computer Vision and Pattern Recognition}, pp. \bibinfo{pages}{3354--3361}.
%Type = Inproceedings
\bibitem[{Godard et~al.(2017)Godard, Mac~Aodha and
  Brostow}]{godard2017unsupervised}
\bibinfo{author}{Godard, C.}, \bibinfo{author}{Mac~Aodha, O.},
  \bibinfo{author}{Brostow, G.J.}, \bibinfo{year}{2017}.
\newblock \bibinfo{title}{Unsupervised monocular depth estimation with
  left-right consistency}, in: \bibinfo{booktitle}{Proceedings of the IEEE
  Conference on Computer Vision and Pattern Recognition}, pp.
  \bibinfo{pages}{270--279}.
%Type = Inproceedings
\bibitem[{Godard et~al.(2019)Godard, Mac~Aodha, Firman and
  Brostow}]{godard2019digging}
\bibinfo{author}{Godard, C.}, \bibinfo{author}{Mac~Aodha, O.},
  \bibinfo{author}{Firman, M.}, \bibinfo{author}{Brostow, G.J.},
  \bibinfo{year}{2019}.
\newblock \bibinfo{title}{Digging into self-supervised monocular depth
  estimation}, in: \bibinfo{booktitle}{Proceedings of the IEEE International
  Conference on Computer Vision}, pp. \bibinfo{pages}{3828--3838}.
%Type = Article
\bibitem[{Goodfellow et~al.(2014)Goodfellow, Pouget-Abadie, Mirza, Xu,
  Warde-Farley, Ozair, Courville and Bengio}]{goodfellow2014generative}
\bibinfo{author}{Goodfellow, I.J.}, \bibinfo{author}{Pouget-Abadie, J.},
  \bibinfo{author}{Mirza, M.}, \bibinfo{author}{Xu, B.},
  \bibinfo{author}{Warde-Farley, D.}, \bibinfo{author}{Ozair, S.},
  \bibinfo{author}{Courville, A.}, \bibinfo{author}{Bengio, Y.},
  \bibinfo{year}{2014}.
\newblock \bibinfo{title}{Generative adversarial networks}.
\newblock \bibinfo{journal}{arXiv preprint arXiv:1406.2661} .
%Type = Inproceedings
\bibitem[{He et~al.(2016)He, Zhang, Ren and Sun}]{he2016deep}
\bibinfo{author}{He, K.}, \bibinfo{author}{Zhang, X.}, \bibinfo{author}{Ren,
  S.}, \bibinfo{author}{Sun, J.}, \bibinfo{year}{2016}.
\newblock \bibinfo{title}{Deep residual learning for image recognition}, in:
  \bibinfo{booktitle}{Proceedings of the IEEE Conference on Computer Vision and
  Pattern Recognition}, pp. \bibinfo{pages}{770--778}.
%Type = Article
\bibitem[{He et~al.(2018)He, Wang and Hu}]{he2018learning}
\bibinfo{author}{He, L.}, \bibinfo{author}{Wang, G.}, \bibinfo{author}{Hu, Z.},
  \bibinfo{year}{2018}.
\newblock \bibinfo{title}{Learning depth from single images with deep neural
  network embedding focal length}.
\newblock \bibinfo{journal}{IEEE Transactions on Image Processing}
  \bibinfo{volume}{27}, \bibinfo{pages}{4676--4689}.
%Type = Article
\bibitem[{Horn and Schunck(1981)}]{horn1981determining}
\bibinfo{author}{Horn, B.K.}, \bibinfo{author}{Schunck, B.G.},
  \bibinfo{year}{1981}.
\newblock \bibinfo{title}{Determining optical flow}.
\newblock \bibinfo{journal}{Artificial intelligence} \bibinfo{volume}{17},
  \bibinfo{pages}{185--203}.
%Type = Inproceedings
\bibitem[{Hur and Roth(2019)}]{hur2019iterative}
\bibinfo{author}{Hur, J.}, \bibinfo{author}{Roth, S.}, \bibinfo{year}{2019}.
\newblock \bibinfo{title}{Iterative residual refinement for joint optical flow
  and occlusion estimation}, in: \bibinfo{booktitle}{Proceedings of the
  IEEE/CVF Conference on Computer Vision and Pattern Recognition}, pp.
  \bibinfo{pages}{5754--5763}.
%Type = Inproceedings
\bibitem[{Jaderberg et~al.(2015)Jaderberg, Simonyan, Zisserman
  et~al.}]{jaderberg2015spatial}
\bibinfo{author}{Jaderberg, M.}, \bibinfo{author}{Simonyan, K.},
  \bibinfo{author}{Zisserman, A.}, et~al., \bibinfo{year}{2015}.
\newblock \bibinfo{title}{Spatial transformer networks}, in:
  \bibinfo{booktitle}{Advances in Neural Information Processing Systems}, pp.
  \bibinfo{pages}{2017--2025}.
%Type = Inproceedings
\bibitem[{Johnston and Carneiro(2020)}]{johnston2020self}
\bibinfo{author}{Johnston, A.}, \bibinfo{author}{Carneiro, G.},
  \bibinfo{year}{2020}.
\newblock \bibinfo{title}{Self-supervised monocular trained depth estimation
  using self-attention and discrete disparity volume}, in:
  \bibinfo{booktitle}{Proceedings of the IEEE/CVF Conference on Computer Vision
  and Pattern Recognition}, pp. \bibinfo{pages}{4756--4765}.
%Type = Inproceedings
\bibitem[{Kharbat et~al.(2008)Kharbat, Aouf, Tsourdos and
  White}]{kharbat2008robust}
\bibinfo{author}{Kharbat, M.}, \bibinfo{author}{Aouf, N.},
  \bibinfo{author}{Tsourdos, A.}, \bibinfo{author}{White, B.A.},
  \bibinfo{year}{2008}.
\newblock \bibinfo{title}{Robust brightness description for computing optical
  flow.}, in: \bibinfo{booktitle}{BMVC}, \bibinfo{organization}{Citeseer}. pp.
  \bibinfo{pages}{1--10}.
%Type = Article
\bibitem[{Kingma and Ba(2014)}]{kingma2014adam}
\bibinfo{author}{Kingma, D.P.}, \bibinfo{author}{Ba, J.}, \bibinfo{year}{2014}.
\newblock \bibinfo{title}{Adam: A method for stochastic optimization}.
\newblock \bibinfo{journal}{arXiv preprint arXiv:1412.6980} .
%Type = Article
\bibitem[{Leonard et~al.(2018)Leonard, Sinha, Reiter, Ishii, Gallia, Taylor and
  Hager}]{leonard2018evaluation}
\bibinfo{author}{Leonard, S.}, \bibinfo{author}{Sinha, A.},
  \bibinfo{author}{Reiter, A.}, \bibinfo{author}{Ishii, M.},
  \bibinfo{author}{Gallia, G.L.}, \bibinfo{author}{Taylor, R.H.},
  \bibinfo{author}{Hager, G.D.}, \bibinfo{year}{2018}.
\newblock \bibinfo{title}{Evaluation and stability analysis of video-based
  navigation system for functional endoscopic sinus surgery on in vivo clinical
  data}.
\newblock \bibinfo{journal}{IEEE transactions on medical imaging}
  \bibinfo{volume}{37}, \bibinfo{pages}{2185--2195}.
%Type = Article
\bibitem[{Li et~al.(2020)Li, Li, Yang, Ding, Jolfaei and
  Zheng}]{li2020unsupervised}
\bibinfo{author}{Li, L.}, \bibinfo{author}{Li, X.}, \bibinfo{author}{Yang, S.},
  \bibinfo{author}{Ding, S.}, \bibinfo{author}{Jolfaei, A.},
  \bibinfo{author}{Zheng, X.}, \bibinfo{year}{2020}.
\newblock \bibinfo{title}{Unsupervised learning-based continuous depth and
  motion estimation with monocular endoscopy for virtual reality minimally
  invasive surgery}.
\newblock \bibinfo{journal}{IEEE Transactions on Industrial Informatics} .
%Type = Article
\bibitem[{Liu et~al.(2015)Liu, Shen, Lin and Reid}]{liu2015learning}
\bibinfo{author}{Liu, F.}, \bibinfo{author}{Shen, C.}, \bibinfo{author}{Lin,
  G.}, \bibinfo{author}{Reid, I.}, \bibinfo{year}{2015}.
\newblock \bibinfo{title}{Learning depth from single monocular images using
  deep convolutional neural fields}.
\newblock \bibinfo{journal}{IEEE transactions on pattern analysis and machine
  intelligence} \bibinfo{volume}{38}, \bibinfo{pages}{2024--2039}.
%Type = Article
\bibitem[{Liu et~al.(2019)Liu, Sinha, Ishii, Hager, Reiter, Taylor and
  Unberath}]{liu2019dense}
\bibinfo{author}{Liu, X.}, \bibinfo{author}{Sinha, A.}, \bibinfo{author}{Ishii,
  M.}, \bibinfo{author}{Hager, G.D.}, \bibinfo{author}{Reiter, A.},
  \bibinfo{author}{Taylor, R.H.}, \bibinfo{author}{Unberath, M.},
  \bibinfo{year}{2019}.
\newblock \bibinfo{title}{Dense depth estimation in monocular endoscopy with
  self-supervised learning methods}.
\newblock \bibinfo{journal}{IEEE Transactions on Medical Imaging}
  \bibinfo{volume}{39}, \bibinfo{pages}{1438--1447}.
%Type = Article
\bibitem[{Lorensen and Cline(1987)}]{lorensen1987marching}
\bibinfo{author}{Lorensen, W.E.}, \bibinfo{author}{Cline, H.E.},
  \bibinfo{year}{1987}.
\newblock \bibinfo{title}{Marching cubes: A high resolution 3d surface
  construction algorithm}.
\newblock \bibinfo{journal}{ACM siggraph computer graphics}
  \bibinfo{volume}{21}, \bibinfo{pages}{163--169}.
%Type = Article
\bibitem[{Lowe(2004)}]{lowe2004distinctive}
\bibinfo{author}{Lowe, D.G.}, \bibinfo{year}{2004}.
\newblock \bibinfo{title}{Distinctive image features from scale-invariant
  keypoints}.
\newblock \bibinfo{journal}{International journal of computer vision}
  \bibinfo{volume}{60}, \bibinfo{pages}{91--110}.
%Type = Article
\bibitem[{Luo et~al.(2019)Luo, Yang, Wang, Wang, Xu, Nevatia and
  Yuille}]{luo2019every}
\bibinfo{author}{Luo, C.}, \bibinfo{author}{Yang, Z.}, \bibinfo{author}{Wang,
  P.}, \bibinfo{author}{Wang, Y.}, \bibinfo{author}{Xu, W.},
  \bibinfo{author}{Nevatia, R.}, \bibinfo{author}{Yuille, A.},
  \bibinfo{year}{2019}.
\newblock \bibinfo{title}{Every pixel counts++: Joint learning of geometry and
  motion with 3d holistic understanding.}
\newblock \bibinfo{journal}{IEEE Transactions on Pattern Analysis and Machine
  Intelligence} .
%Type = Inproceedings
\bibitem[{Mahjourian et~al.(2018)Mahjourian, Wicke and
  Angelova}]{mahjourian2018unsupervised}
\bibinfo{author}{Mahjourian, R.}, \bibinfo{author}{Wicke, M.},
  \bibinfo{author}{Angelova, A.}, \bibinfo{year}{2018}.
\newblock \bibinfo{title}{Unsupervised learning of depth and ego-motion from
  monocular video using 3d geometric constraints}, in:
  \bibinfo{booktitle}{Proceedings of the IEEE Conference on Computer Vision and
  Pattern Recognition}, pp. \bibinfo{pages}{5667--5675}.
%Type = Article
\bibitem[{Mahmood et~al.(2018)Mahmood, Chen and Durr}]{mahmood2018unsupervised}
\bibinfo{author}{Mahmood, F.}, \bibinfo{author}{Chen, R.},
  \bibinfo{author}{Durr, N.J.}, \bibinfo{year}{2018}.
\newblock \bibinfo{title}{Unsupervised reverse domain adaptation for synthetic
  medical images via adversarial training}.
\newblock \bibinfo{journal}{IEEE transactions on medical imaging}
  \bibinfo{volume}{37}, \bibinfo{pages}{2572--2581}.
%Type = Article
\bibitem[{Mahmood and Durr(2018)}]{mahmood2018deep}
\bibinfo{author}{Mahmood, F.}, \bibinfo{author}{Durr, N.J.},
  \bibinfo{year}{2018}.
\newblock \bibinfo{title}{Deep learning and conditional random fields-based
  depth estimation and topographical reconstruction from conventional
  endoscopy}.
\newblock \bibinfo{journal}{Medical Image Analysis} \bibinfo{volume}{48},
  \bibinfo{pages}{230--243}.
%Type = Article
\bibitem[{Mur-Artal et~al.(2015)Mur-Artal, Montiel and Tardos}]{mur2015orb}
\bibinfo{author}{Mur-Artal, R.}, \bibinfo{author}{Montiel, J.M.M.},
  \bibinfo{author}{Tardos, J.D.}, \bibinfo{year}{2015}.
\newblock \bibinfo{title}{Orb-slam: a versatile and accurate monocular slam
  system}.
\newblock \bibinfo{journal}{IEEE transactions on robotics}
  \bibinfo{volume}{31}, \bibinfo{pages}{1147--1163}.
%Type = Article
\bibitem[{Ozyoruk et~al.(2021)Ozyoruk, Gokceler, Bobrow, Coskun, Incetan,
  Almalioglu, Mahmood, Curto, Perdigoto, Oliveira
  et~al.}]{bengisu2020quantitative}
\bibinfo{author}{Ozyoruk, K.B.}, \bibinfo{author}{Gokceler, G.I.},
  \bibinfo{author}{Bobrow, T.L.}, \bibinfo{author}{Coskun, G.},
  \bibinfo{author}{Incetan, K.}, \bibinfo{author}{Almalioglu, Y.},
  \bibinfo{author}{Mahmood, F.}, \bibinfo{author}{Curto, E.},
  \bibinfo{author}{Perdigoto, L.}, \bibinfo{author}{Oliveira, M.}, et~al.,
  \bibinfo{year}{2021}.
\newblock \bibinfo{title}{Endoslam dataset and an unsupervised monocular visual
  odometry and depth estimation approach for endoscopic videos:
  Endo-sfmlearner}.
\newblock \bibinfo{journal}{Medical Image Analysis} .
%Type = Article
\bibitem[{Paszke et~al.(2017)Paszke, Gross, Chintala, Chanan, Yang, DeVito,
  Lin, Desmaison, Antiga and Lerer}]{paszke2017automatic}
\bibinfo{author}{Paszke, A.}, \bibinfo{author}{Gross, S.},
  \bibinfo{author}{Chintala, S.}, \bibinfo{author}{Chanan, G.},
  \bibinfo{author}{Yang, E.}, \bibinfo{author}{DeVito, Z.},
  \bibinfo{author}{Lin, Z.}, \bibinfo{author}{Desmaison, A.},
  \bibinfo{author}{Antiga, L.}, \bibinfo{author}{Lerer, A.},
  \bibinfo{year}{2017}.
\newblock \bibinfo{title}{Automatic differentiation in pytorch} .
%Type = Inproceedings
\bibitem[{Ranjan et~al.(2019)Ranjan, Jampani, Balles, Kim, Sun, Wulff and
  Black}]{ranjan2019competitive}
\bibinfo{author}{Ranjan, A.}, \bibinfo{author}{Jampani, V.},
  \bibinfo{author}{Balles, L.}, \bibinfo{author}{Kim, K.},
  \bibinfo{author}{Sun, D.}, \bibinfo{author}{Wulff, J.},
  \bibinfo{author}{Black, M.J.}, \bibinfo{year}{2019}.
\newblock \bibinfo{title}{Competitive collaboration: Joint unsupervised
  learning of depth, camera motion, optical flow and motion segmentation}, in:
  \bibinfo{booktitle}{Proceedings of the IEEE Conference on Computer Vision and
  Pattern Recognition}, pp. \bibinfo{pages}{12240--12249}.
%Type = Article
\bibitem[{Recasens et~al.(2021)Recasens, Lamarca, F{\'a}cil, Montiel and
  Civera}]{recasens2021endo}
\bibinfo{author}{Recasens, D.}, \bibinfo{author}{Lamarca, J.},
  \bibinfo{author}{F{\'a}cil, J.M.}, \bibinfo{author}{Montiel, J.},
  \bibinfo{author}{Civera, J.}, \bibinfo{year}{2021}.
\newblock \bibinfo{title}{Endo-depth-and-motion: Reconstruction and tracking in
  endoscopic videos using depth networks and photometric constraints}.
\newblock \bibinfo{journal}{IEEE Robotics and Automation Letters}
  \bibinfo{volume}{6}, \bibinfo{pages}{7225--7232}.
%Type = Inproceedings
\bibitem[{Ren et~al.(2017)Ren, He, Peng, Liu, Zhu and Zeng}]{ren2017shape}
\bibinfo{author}{Ren, Z.}, \bibinfo{author}{He, T.}, \bibinfo{author}{Peng,
  L.}, \bibinfo{author}{Liu, S.}, \bibinfo{author}{Zhu, S.},
  \bibinfo{author}{Zeng, B.}, \bibinfo{year}{2017}.
\newblock \bibinfo{title}{Shape recovery of endoscopic videos by shape from
  shading using mesh regularization}, in: \bibinfo{booktitle}{International
  Conference on Image and Graphics}, \bibinfo{organization}{Springer}. pp.
  \bibinfo{pages}{204--213}.
%Type = Inproceedings
\bibitem[{Repala and Dubey(2019)}]{repala2019dual}
\bibinfo{author}{Repala, V.K.}, \bibinfo{author}{Dubey, S.R.},
  \bibinfo{year}{2019}.
\newblock \bibinfo{title}{Dual cnn models for unsupervised monocular depth
  estimation}, in: \bibinfo{booktitle}{International Conference on Pattern
  Recognition and Machine Intelligence}, \bibinfo{organization}{Springer}. pp.
  \bibinfo{pages}{209--217}.
%Type = Article
\bibitem[{Saxena et~al.(2008)Saxena, Sun and Ng}]{saxena2008make3d}
\bibinfo{author}{Saxena, A.}, \bibinfo{author}{Sun, M.}, \bibinfo{author}{Ng,
  A.Y.}, \bibinfo{year}{2008}.
\newblock \bibinfo{title}{Make3d: Learning 3d scene structure from a single
  still image}.
\newblock \bibinfo{journal}{IEEE transactions on pattern analysis and machine
  intelligence} \bibinfo{volume}{31}, \bibinfo{pages}{824--840}.
%Type = Article
\bibitem[{Shao et~al.(2021a)Shao, Li, Pei, Liu, Chen, Zhu, Wu and
  Zhang}]{shao2021nenet}
\bibinfo{author}{Shao, S.}, \bibinfo{author}{Li, R.}, \bibinfo{author}{Pei,
  Z.}, \bibinfo{author}{Liu, Z.}, \bibinfo{author}{Chen, W.},
  \bibinfo{author}{Zhu, W.}, \bibinfo{author}{Wu, X.}, \bibinfo{author}{Zhang,
  B.}, \bibinfo{year}{2021}a.
\newblock \bibinfo{title}{Nenet: Monocular depth estimation via neural
  ensembles}.
\newblock \bibinfo{journal}{arXiv preprint arXiv:2111.08313} .
%Type = Inproceedings
\bibitem[{Shao et~al.(2021b)Shao, Pei, Chen, Zhang, Wu, Sun and
  Doermann}]{shao2021self}
\bibinfo{author}{Shao, S.}, \bibinfo{author}{Pei, Z.}, \bibinfo{author}{Chen,
  W.}, \bibinfo{author}{Zhang, B.}, \bibinfo{author}{Wu, X.},
  \bibinfo{author}{Sun, D.}, \bibinfo{author}{Doermann, D.},
  \bibinfo{year}{2021}b.
\newblock \bibinfo{title}{Self-supervised learning for monocular depth
  estimation on minimally invasive surgery scenes}, in:
  \bibinfo{booktitle}{2021 IEEE International Conference on Robotics and
  Automation (ICRA)}, \bibinfo{organization}{IEEE}. pp.
  \bibinfo{pages}{7159--7165}.
%Type = Article
\bibitem[{Shao et~al.(2021c)Shao, Pei, Chen, Zhu, Wu and Zhang}]{shao2021multi}
\bibinfo{author}{Shao, S.}, \bibinfo{author}{Pei, Z.}, \bibinfo{author}{Chen,
  W.}, \bibinfo{author}{Zhu, W.}, \bibinfo{author}{Wu, X.},
  \bibinfo{author}{Zhang, B.}, \bibinfo{year}{2021}c.
\newblock \bibinfo{title}{A multi-scale unsupervised learning for deformable
  image registration}.
\newblock \bibinfo{journal}{International Journal of Computer Assisted
  Radiology and Surgery} , \bibinfo{pages}{1--10}.
%Type = Inproceedings
\bibitem[{Shu et~al.(2020)Shu, Yu, Duan and Yang}]{shu2020feature}
\bibinfo{author}{Shu, C.}, \bibinfo{author}{Yu, K.}, \bibinfo{author}{Duan,
  Z.}, \bibinfo{author}{Yang, K.}, \bibinfo{year}{2020}.
\newblock \bibinfo{title}{Feature-metric loss for self-supervised learning of
  depth and egomotion}, in: \bibinfo{booktitle}{European Conference on Computer
  Vision}, \bibinfo{organization}{Springer}. pp. \bibinfo{pages}{572--588}.
%Type = Inproceedings
\bibitem[{Spencer et~al.(2020)Spencer, Bowden and Hadfield}]{spencer2020defeat}
\bibinfo{author}{Spencer, J.}, \bibinfo{author}{Bowden, R.},
  \bibinfo{author}{Hadfield, S.}, \bibinfo{year}{2020}.
\newblock \bibinfo{title}{Defeat-net: General monocular depth via simultaneous
  unsupervised representation learning}, in: \bibinfo{booktitle}{Proceedings of
  the IEEE Conference on Computer Vision and Pattern Recognition}, pp.
  \bibinfo{pages}{14402--14413}.
%Type = Inproceedings
\bibitem[{Sun et~al.(2019)Sun, Yao, Zhou and Zhao}]{Sun_2019_CVPR}
\bibinfo{author}{Sun, D.}, \bibinfo{author}{Yao, A.}, \bibinfo{author}{Zhou,
  A.}, \bibinfo{author}{Zhao, H.}, \bibinfo{year}{2019}.
\newblock \bibinfo{title}{Deeply-supervised knowledge synergy}, in:
  \bibinfo{booktitle}{Proceedings of the IEEE/CVF Conference on Computer Vision
  and Pattern Recognition (CVPR)}.
%Type = Inproceedings
\bibitem[{Teed and Deng(2020)}]{teed2020raft}
\bibinfo{author}{Teed, Z.}, \bibinfo{author}{Deng, J.}, \bibinfo{year}{2020}.
\newblock \bibinfo{title}{Raft: Recurrent all-pairs field transforms for
  optical flow}, in: \bibinfo{booktitle}{European Conference on Computer
  Vision}, \bibinfo{organization}{Springer}. pp. \bibinfo{pages}{402--419}.
%Type = Inproceedings
\bibitem[{Turan et~al.(2018)Turan, Ornek, Ibrahimli, Giracoglu, Almalioglu,
  Yanik and Sitti}]{turan2018unsupervised}
\bibinfo{author}{Turan, M.}, \bibinfo{author}{Ornek, E.P.},
  \bibinfo{author}{Ibrahimli, N.}, \bibinfo{author}{Giracoglu, C.},
  \bibinfo{author}{Almalioglu, Y.}, \bibinfo{author}{Yanik, M.F.},
  \bibinfo{author}{Sitti, M.}, \bibinfo{year}{2018}.
\newblock \bibinfo{title}{Unsupervised odometry and depth learning for
  endoscopic capsule robots}, in: \bibinfo{booktitle}{2018 IEEE/RSJ
  International Conference on Intelligent Robots and Systems (IROS)},
  \bibinfo{organization}{IEEE}. pp. \bibinfo{pages}{1801--1807}.
%Type = Article
\bibitem[{Visentini-Scarzanella et~al.(2017)Visentini-Scarzanella, Sugiura,
  Kaneko and Koto}]{visentini2017deep}
\bibinfo{author}{Visentini-Scarzanella, M.}, \bibinfo{author}{Sugiura, T.},
  \bibinfo{author}{Kaneko, T.}, \bibinfo{author}{Koto, S.},
  \bibinfo{year}{2017}.
\newblock \bibinfo{title}{Deep monocular 3d reconstruction for assisted
  navigation in bronchoscopy}.
\newblock \bibinfo{journal}{International Journal of Computer Assisted
  Radiology and Surgery} \bibinfo{volume}{12}, \bibinfo{pages}{1089--1099}.
%Type = Inproceedings
\bibitem[{Wang et~al.(2018)Wang, Yang, Yang, Zhao, Wang and
  Xu}]{wang2018occlusion}
\bibinfo{author}{Wang, Y.}, \bibinfo{author}{Yang, Y.}, \bibinfo{author}{Yang,
  Z.}, \bibinfo{author}{Zhao, L.}, \bibinfo{author}{Wang, P.},
  \bibinfo{author}{Xu, W.}, \bibinfo{year}{2018}.
\newblock \bibinfo{title}{Occlusion aware unsupervised learning of optical
  flow}, in: \bibinfo{booktitle}{Proceedings of the IEEE Conference on Computer
  Vision and Pattern Recognition}, pp. \bibinfo{pages}{4884--4893}.
%Type = Article
\bibitem[{Wang et~al.(2004)Wang, Bovik, Sheikh and Simoncelli}]{wang2004image}
\bibinfo{author}{Wang, Z.}, \bibinfo{author}{Bovik, A.C.},
  \bibinfo{author}{Sheikh, H.R.}, \bibinfo{author}{Simoncelli, E.P.},
  \bibinfo{year}{2004}.
\newblock \bibinfo{title}{Image quality assessment: from error visibility to
  structural similarity}.
\newblock \bibinfo{journal}{IEEE transactions on image processing}
  \bibinfo{volume}{13}, \bibinfo{pages}{600--612}.
%Type = Inproceedings
\bibitem[{Xu et~al.(2017)Xu, Ricci, Ouyang, Wang and Sebe}]{xu2017multi}
\bibinfo{author}{Xu, D.}, \bibinfo{author}{Ricci, E.}, \bibinfo{author}{Ouyang,
  W.}, \bibinfo{author}{Wang, X.}, \bibinfo{author}{Sebe, N.},
  \bibinfo{year}{2017}.
\newblock \bibinfo{title}{Multi-scale continuous crfs as sequential deep
  networks for monocular depth estimation}, in: \bibinfo{booktitle}{Proceedings
  of the IEEE Conference on Computer Vision and Pattern Recognition}, pp.
  \bibinfo{pages}{5354--5362}.
%Type = Inproceedings
\bibitem[{Xu et~al.(2018)Xu, Wang, Tang, Liu, Sebe and
  Ricci}]{xu2018structured}
\bibinfo{author}{Xu, D.}, \bibinfo{author}{Wang, W.}, \bibinfo{author}{Tang,
  H.}, \bibinfo{author}{Liu, H.}, \bibinfo{author}{Sebe, N.},
  \bibinfo{author}{Ricci, E.}, \bibinfo{year}{2018}.
\newblock \bibinfo{title}{Structured attention guided convolutional neural
  fields for monocular depth estimation}, in: \bibinfo{booktitle}{Proceedings
  of the IEEE conference on computer vision and pattern recognition}, pp.
  \bibinfo{pages}{3917--3925}.
%Type = Inproceedings
\bibitem[{Yang et~al.(2020)Yang, Stumberg, Wang and Cremers}]{yang2020d3vo}
\bibinfo{author}{Yang, N.}, \bibinfo{author}{Stumberg, L.v.},
  \bibinfo{author}{Wang, R.}, \bibinfo{author}{Cremers, D.},
  \bibinfo{year}{2020}.
\newblock \bibinfo{title}{D3vo: Deep depth, deep pose and deep uncertainty for
  monocular visual odometry}, in: \bibinfo{booktitle}{Proceedings of the IEEE
  Conference on Computer Vision and Pattern Recognition}, pp.
  \bibinfo{pages}{1281--1292}.
%Type = Inproceedings
\bibitem[{Yang et~al.(2018)Yang, Wang, Wang, Xu and Nevatia}]{yang2018lego}
\bibinfo{author}{Yang, Z.}, \bibinfo{author}{Wang, P.}, \bibinfo{author}{Wang,
  Y.}, \bibinfo{author}{Xu, W.}, \bibinfo{author}{Nevatia, R.},
  \bibinfo{year}{2018}.
\newblock \bibinfo{title}{Lego: Learning edge with geometry all at once by
  watching videos}, in: \bibinfo{booktitle}{Proceedings of the IEEE conference
  on computer vision and pattern recognition}, pp. \bibinfo{pages}{225--234}.
%Type = Inproceedings
\bibitem[{Yin and Shi(2018)}]{yin2018geonet}
\bibinfo{author}{Yin, Z.}, \bibinfo{author}{Shi, J.}, \bibinfo{year}{2018}.
\newblock \bibinfo{title}{Geonet: Unsupervised learning of dense depth, optical
  flow and camera pose}, in: \bibinfo{booktitle}{Proceedings of the IEEE
  Conference on Computer Vision and Pattern Recognition}, pp.
  \bibinfo{pages}{1983--1992}.
%Type = Inproceedings
\bibitem[{Zhan et~al.(2018)Zhan, Garg, Saroj~Weerasekera, Li, Agarwal and
  Reid}]{zhan2018unsupervised}
\bibinfo{author}{Zhan, H.}, \bibinfo{author}{Garg, R.},
  \bibinfo{author}{Saroj~Weerasekera, C.}, \bibinfo{author}{Li, K.},
  \bibinfo{author}{Agarwal, H.}, \bibinfo{author}{Reid, I.},
  \bibinfo{year}{2018}.
\newblock \bibinfo{title}{Unsupervised learning of monocular depth estimation
  and visual odometry with deep feature reconstruction}, in:
  \bibinfo{booktitle}{Proceedings of the IEEE Conference on Computer Vision and
  Pattern Recognition}, pp. \bibinfo{pages}{340--349}.
%Type = Inproceedings
\bibitem[{Zhao et~al.(2017)Zhao, Shi, Qi, Wang and Jia}]{zhao2017pyramid}
\bibinfo{author}{Zhao, H.}, \bibinfo{author}{Shi, J.}, \bibinfo{author}{Qi,
  X.}, \bibinfo{author}{Wang, X.}, \bibinfo{author}{Jia, J.},
  \bibinfo{year}{2017}.
\newblock \bibinfo{title}{Pyramid scene parsing network}, in:
  \bibinfo{booktitle}{Proceedings of the IEEE conference on computer vision and
  pattern recognition}, pp. \bibinfo{pages}{2881--2890}.
%Type = Inproceedings
\bibitem[{Zhou et~al.(2019)Zhou, Wang, Qin and Zeng}]{zhou2019unsupervised}
\bibinfo{author}{Zhou, J.}, \bibinfo{author}{Wang, Y.}, \bibinfo{author}{Qin,
  K.}, \bibinfo{author}{Zeng, W.}, \bibinfo{year}{2019}.
\newblock \bibinfo{title}{Unsupervised high-resolution depth learning from
  videos with dual networks}, in: \bibinfo{booktitle}{Proceedings of the
  IEEE/CVF International Conference on Computer Vision}, pp.
  \bibinfo{pages}{6872--6881}.
%Type = Inproceedings
\bibitem[{Zhou et~al.(2017)Zhou, Brown, Snavely and
  Lowe}]{zhou2017unsupervised}
\bibinfo{author}{Zhou, T.}, \bibinfo{author}{Brown, M.},
  \bibinfo{author}{Snavely, N.}, \bibinfo{author}{Lowe, D.G.},
  \bibinfo{year}{2017}.
\newblock \bibinfo{title}{Unsupervised learning of depth and ego-motion from
  video}, in: \bibinfo{booktitle}{Proceedings of the IEEE conference on
  computer vision and pattern recognition}, pp. \bibinfo{pages}{1851--1858}.
%Type = Misc
\bibitem[{Zhu et~al.(2021)Zhu, Huang, Xu, Qian, Fan and Xie}]{zhu2021test}
\bibinfo{author}{Zhu, W.}, \bibinfo{author}{Huang, Y.}, \bibinfo{author}{Xu,
  D.}, \bibinfo{author}{Qian, Z.}, \bibinfo{author}{Fan, W.},
  \bibinfo{author}{Xie, X.}, \bibinfo{year}{2021}.
\newblock \bibinfo{title}{Test-time training for deformable multi-scale image
  registration}.
%Type = Inproceedings
\bibitem[{Zhu et~al.(2017)Zhu, Xiong, Dai, Yuan and Wei}]{zhu2017deep}
\bibinfo{author}{Zhu, X.}, \bibinfo{author}{Xiong, Y.}, \bibinfo{author}{Dai,
  J.}, \bibinfo{author}{Yuan, L.}, \bibinfo{author}{Wei, Y.},
  \bibinfo{year}{2017}.
\newblock \bibinfo{title}{Deep feature flow for video recognition}, in:
  \bibinfo{booktitle}{Proceedings of the IEEE conference on computer vision and
  pattern recognition}, pp. \bibinfo{pages}{2349--2358}.

\end{thebibliography}
\end{document}